\documentclass{elsarticle_arxiv}

\biboptions{numbers, authoryear, round, colon} 
\usepackage{bm}
\usepackage{amsmath}
\usepackage{amssymb}
\usepackage{dsfont}
\usepackage{stackengine}
\usepackage[export]{adjustbox}
\usepackage{multicol}
\usepackage{floatrow}

\begin{document}
\begin{frontmatter}
\title{Classification and clustering for observations of event time data using non-homogeneous Poisson process models}
\author[rvt, cor1]{Duncan S Barrack}
\ead{duncan.barrack@googlemail.com}
\author[focal]{Simon Preston}
\cortext[cor1]{Corresponding author}
\address[rvt]{Horizon Digital Economy Research Institute, University of Nottingham, Nottingham, UK.}
\address[focal]{School of Mathematical Sciences, University of Nottingham, Nottingham, UK.}

\begin{abstract}
Data of the form of event times arise in various applications. A simple model for such data is a non-homogeneous Poisson process (NHPP) which is specified by a rate function that depends on time. We consider the problem of having access to multiple independent observations of event time data, observed on a common interval, from which we wish to classify or cluster the observations according to their rate functions. Each rate function is unknown but assumed to belong to a finite number of rate functions each defining a distinct class. We model the rate functions using a spline basis expansion, the coefficients of which need to be estimated from data. The classification approach consists of using training data for which the class membership is known, to calculate maximum likelihood estimates of the coefficients for each group, then assigning test observations to a group by a maximum likelihood criterion. For clustering, by analogy to the Gaussian mixture model approach for Euclidean data, we consider mixtures of NHPP and use the expectation-maximisation algorithm to estimate the coefficients of the rate functions for the component models and group membership probabilities for each observation. The classification and clustering approaches perform well on both synthetic and real-world data sets. Code associated with this paper is available at https://github.com/duncan-barrack/NHPP.

\end{abstract}

\end{frontmatter}

\section{Introduction}
\label{intro}
Much real world data is amenable to modelling as a non-homogeneous Poisson process (NHPP) (e.g. stock purchase transaction times \citep{engle2000econometrics}, email traffic \citep{malmgren2008poissonian} as well as computer network traffic \citep{paxson1995wide}). In various applications a natural unit of observation is a sample of event times observed on some common interval $\{t: 0<t \leq T \}$  and a challenge is to label each observation as belonging to one of a small number of distinct classes. A particular example we consider later is the categorisation of stores by till transaction times. This can provide retailers valuable insights in to the different demands placed on different store types and help to inform various decisions from stock management to appropriate store staffing levels. Another example we consider is the categorisation of transportation hubs according to the times at which they are used. Such information helps inform the type and number of services (transportation, catering, retail and others) that should be made available at each type of hub. When a `training set', containing observations with known class memberships is available, such a task is termed `classification', when no labels are available for the data it is termed a `clustering' problem \citep{hastie2005elements}. Previous approaches to classification and clustering of event time data have focused on deriving summary statistics from the data such as the mean and variance of the inter-event times and using these as inputs in statistical learning models \citep{moore2005internet, roughan2004class, mcgregor2004flow}. In addition to studies of temporal data a number of authors have considered spatial point processes. Illian \emph{et al} used functional principal component analysis on second order summary statistics of spatial point process data to classify plant species  \citep{illian2006principal, illian2004multivariate}. Closer in approach to the present work \citet{linnett1995texture} used a maximum likelihood (ML) criterion to classify the textures of images modelled using homogeneous Poisson processes (HPPs). 

HPPs, like those used in the work of \citet{linnett1995texture} assume rates are constant for each class which is too restrictive for many applications. For example stock transaction times and human communication behaviour both exhibit periodic patterns \citep{engle2000econometrics, malmgren2008poissonian}. Computer network traffic is better modelled by non-homogeneous NHPP models rather than HPPs \citep{paxson1995wide}. Hence, in this work, we consider instead NHPP models, which are defined in terms of rate functions that vary with time. We model these rate functions using a spline basis expansion which imposes an assumption of smoothness on the functions and means estimating them amounts to estimating a finite number of basis coefficients. We go beyond previous work which has focused purely on rate function estimation \citep{alizadeh2008arrival, kuhl2000new, scott1980nonparametric, zhao1996maximum, massey1996estimating, lee1991modeling} by considering the tasks of classification and clustering. For classification we show that once intensity estimates for the NHPP rate functions corresponding to each class are obtained by solving a convex optmisation problem, it is a simple task to obtain the posterior class membership probabilities for test data. For clustering, where class labels are not known \emph{a priori}, we assume all data to have been generated from a mixture of $k$ NHPP models. Again the rate functions for each model are represented by a combination of basis functions but his time the expectation-maximisation (EM) algorithm is used to obtain the coefficients of the basis functions as well as $k$ membership probabilities for each distinct observation of event times. Our approaches are validated on synthetic and three real data sets.

The paper is structured as follows. In Section \ref{sec:rate_func} we describe our procedure for estimating NHPP rate functions. Then in Sections \ref{sec:classification} and \ref{sec:clustering} we outline our classification and clustering methods. The results of applying these to synthetic and real data are shown in Sections \ref{sec:data}. We conclude with a discussion in Section \ref{sec:conclusion}.

\section{Rate function estimation}\label{sec:rate_func}
Suppose that we assume that a sample of event times $\boldsymbol{\chi}=\{\chi_{1}, \chi_{2},\ldots, \chi_{m_{\chi}}\}$, observed on the interval $\{t: 0<t \leq T \}$, arises from a NHPP with rate function $\lambda(t)$, then the log likelihood for the rate function is

\begin{eqnarray}
\label{eq:log_lik}
l(\lambda(t)|\boldsymbol{\chi})=-M(t) + \sum_{i=1}^{m_{\chi}}\mbox{ln}(\lambda(\chi_{i})),
\end{eqnarray}

\noindent where $M(t)=\int_{0}^{t} \lambda(\tilde{t}) d\tilde{t}$ (see for example \citet{thompson2012point}).

In estimating $\lambda(t)$ in equation (\ref{eq:log_lik}) we impose that it belongs to a class of smooth functions by assuming it is a linear combination of smooth basis functions
\begin{eqnarray}
\label{eq:basis_rep}
\lambda(t)=\sum _{m=1}^{n_{b}}c_{m}B_{m}(t).
\end{eqnarray}

In this paper, for the basis functions $B_{1}(t), \ldots, B_{n_{b}}(t)$, we use B-splines \citep{de1978practical} which offer control over smoothness and have finite support. We estimate the basis function coefficients $\boldsymbol{c}=\{c_{1}, c_{2}, \ldots , c_{n_{b}} \}$ by minimising $-l(\lambda(t)|\boldsymbol{\chi})$ i.e. by minimising

\begin{eqnarray}
\label{eq:log_lik_max}
f(\boldsymbol{c})= -l(\lambda(t)|\boldsymbol{\chi}) = M(t) -  \sum_{i=1}^{m_{\chi}} \mbox{ln}\left(\sum _{m=1}^{n_{b}}c_{m}B_{m}(\chi_{i})\right),
\end{eqnarray}

\noindent subject to the constraint that the rate function $\lambda(t)$ is non-negative i.e.

\begin{eqnarray}
\label{eq:const}
g(\boldsymbol{c}) = - \lambda(t)= -\sum _{m=1}^{n_{b}}c_{m}B_{m}(t) \leq 0.
\end{eqnarray}

Minimising (\ref{eq:log_lik_max}) subject to (\ref{eq:const}) is a convex optimisation problem which we show as follows.  First, we derive the gradient of $f(\boldsymbol{c})$

\begin{align}
\label{eq:log_lik_grad}
\bigtriangledown f&=\left(\bigtriangledown_{1}(f),\ldots, \bigtriangledown_{n_{b}}(f) \right),
\end{align}

\noindent where $\bigtriangledown_{j}(f) = \frac{\partial}{\partial c_{j}}M(t) - \sum_{i=1}^{m_{\chi}}\frac{B_{i}(\chi_{i})}{\sum _{m=1}^{n_{b}}c_{m}B_{m}(\chi_{i})}$. The $n_{b}\times n_{b}$ Hessian is

\begin{align}
\label{eq:log_lik_hessian}
\mathbf{H}=\begin{bmatrix}
        H_{1,1}(f) & \ldots & H_{1,n_{b}}(f) \\
        \vdots & \ddots & \vdots\\
        H_{n_{b},1}(f) & \ldots & H_{n_{b}, n_{b}}(f)
     \end{bmatrix},
\end{align}

\noindent where $H_{s,u}(f)=\sum_{i=1}^{m_{\chi}} \frac{B_{s}(\chi_{i})B_{u}(\chi_{i})}{\left(\sum _{m=1}^{n_{b}}c_{m}B_{m}(\chi_{i})\right)^{2}}$\\
\noindent as $\frac{\partial^2}{\partial c_{s}\partial c_{u}}M(t) = \int_{t_{1}}^{t} \left( \frac{\partial^2}{\partial c_{s}\partial c_{u}} \sum _{m=1}^{n_{b}}c_{m}B_{m}(\tilde{t}) \right) d\tilde{t} = 0$  for every $s$ and $u$. 

As we may express $\mathbf{H} = \mathbf{X}^{\prime}\mathbf{X}$, where the $\chi_{m_{\chi}} \times n_{b}$ matrix $\mathbf{X}=$
\begin{align}
\label{eq:X}
\begin{bmatrix}
        \frac{B_{1}(\chi_{1})}{\sum _{m=1}^{n_{b}}c_{m}B_{m}(\chi_{1})} & \ldots & \frac{B_{n_{b}}(\chi_{1})}{\sum _{m=1}^{n_{b}}c_{m}B_{m}(\chi_{1})} \\
        \vdots & \ddots & \vdots\\
       \frac{B_{1}(\chi_{m_{\chi}})}{\sum _{m=1}^{n_{b}}c_{m}B_{m}(\chi_{m_{\chi}})} & \ldots & \frac{B_{n_{b}}(\chi_{m_{\chi}})}{\sum _{m=1}^{n_{b}}c_{m}B_{m}(\chi_{m_{\chi}})}
     \end{bmatrix},
\end{align}
\noindent thus $\mathbf{H}$ is positive semi-definitive as $\boldsymbol{y}^\prime \mathbf{H}\boldsymbol{y} = \boldsymbol{y}^\prime \mathbf{X}^\prime \mathbf{X} \boldsymbol{y} = (\boldsymbol{y} \mathbf{X})^\prime (\mathbf{X} \boldsymbol{y}) = ||\mathbf{X} \boldsymbol{y}||^2 \geq 0$ for every non-zero column vector $\boldsymbol{y}=\{y_{1}, y_{2}, \ldots, y_{n_{b}}\}$ of real numbers. Therefore $f(\boldsymbol{c})$ (equation (\ref{eq:log_lik_max})) is convex in  $\boldsymbol{c}$.

Constraint (\ref{eq:const}) is linear in $\boldsymbol{c}$ and hence convex in $\boldsymbol{c}$. Hence estimating the basis function coefficients of the NHPP rate function $\lambda(t)$ is a convex optimisation problem. 

Furthermore, as $f(\boldsymbol{c})$ and $g(\boldsymbol{c})$ are twice differentiable, the minimisation to perform is straightforward using anyone of several numerical techniques designed for solving non-linear convex optimisation problems where the objective function and inequality constraint are both twice differentiable. For the results in this paper we used the interior point method \citep{byrd1999interior, byrd2000trust, waltz2006interior}.

\section{Classification}\label{sec:classification}
Consider a set of $n_{\text{train}}$ training observations $\boldsymbol{T}_{\text{train}}=$\\
\noindent$\{\{\boldsymbol{\chi}_{1}, g_{1}\}, \{\boldsymbol{\chi}_{2}, g_{2} \}, \ldots, \{\boldsymbol{\chi}_{n_{\text{train}}}, g_{n_{\text{train}}} \}\}$ in which $\boldsymbol{\chi}_{l}=\{\chi_{l,1}, \chi_{l,2},\ldots, \chi_{l,m_{l}}\}$ is a set of of $m_{l}$ event times for the $l^{\text{th}}$ observation and $g_{l}\in{1,2,\ldots,k}$ is its corresponding class label. The classification task consists of using this training data to predict the unobserved class label for a new observation $j$ with event times $\boldsymbol{\chi}_{j}=\{\chi_{j,1}, \chi_{j,2},\ldots, \chi_{j,m_{j}}\}$. Assuming each training observation is independent, the log likelihood estimate for the rate function $\lambda_{\nu}(t)$ for class $\nu$ is

\begin{eqnarray}
&l(\lambda_{\nu}(t)|\boldsymbol{T}_{\text{train}})=\label{eq:log_lik_class}
&\sum_{l=1}^{n_{\text{train}}}\mathds{1}(G_{l}=\nu) \left[-M_{\nu}(t) + \sum_{i=1}^{m_{l}}\mbox{ln}(\lambda_{\nu}(\chi_{l,i}))\right], 
\end{eqnarray}

\noindent where the indicator function $\mathds{1}(G_{l}=\nu) = 1$ if $G_{l} = \nu$ and 0 otherwise. The rate function $\lambda_{\nu}(t)$ is given by 

\begin{eqnarray}
\label{eq:basis_rep_est_class}
\lambda_{\nu}(t)=\sum _{m=1}^{n_{b}}c_{\nu,m}B_{m}(t).
\end{eqnarray}

Estimates for the basis function coefficients of this expression are found via the maximum likelihood estimation outlined in Section \ref{sec:rate_func}.

\hfill\break
\noindent \textbf{Classification of test data}\\
\noindent The posterior probability that a test observation $\boldsymbol{\chi}_{j}$ is a member of class $\nu$ can be derived from Bayes' theorem
\begin{eqnarray}
\label{eq:mp_class}
P(g_{j}=\nu|\boldsymbol{\chi}_{j})=\frac{e^{-\widehat{M}_{\nu}(t)} \prod_{i=1}^{m_{j}}\widehat{\lambda}_{\nu}(\chi_{j,i})}{\sum_{q=1}^{k}e^{-\widehat{M}_{q}(t)} \prod_{i=1}^{m_{j}}\widehat{\lambda}_{q}(\chi_{j,i})}.
\end{eqnarray}

Observations are then assigned to the class for which this probability is maximal, i.e. $\widehat{g}=\underset{\nu}{\mathrm{argmax}}\{P(g_{j}=\nu|\boldsymbol{\chi}_{j}) \}$.

\section{Clustering}\label{sec:clustering}
For the unsupervised learning task of cluster analysis of observations of event time samples we assume the data to have been generated from a mixture of NHPP models. Furthermore, although class memberships are unknown, each observation is associated with a hidden latent variable which specifies which group, or class, it belongs to. Specifically, we assume the event times samples from $n$ observations $\boldsymbol{D}=\{\boldsymbol{\chi}_{1}, \boldsymbol{\chi}_{2}, \ldots, \boldsymbol{\chi}_{n}\}$ are generated from a mixture of $k$ NHPP models and let $\boldsymbol{z}=\{z_{1},z_{2},\ldots , z_{n}\}$ be a set of latent random variables which determine the component from which each observation originates. For the mixture the complete data log likelihood function is

\begin{eqnarray}
\begin{aligned}
l(\boldsymbol{\theta}|\boldsymbol{D}, \boldsymbol{z}) &= \sum_{l=1}^{n} \mbox{ln} \left( p(\boldsymbol{\chi}_{l}, \boldsymbol{z} |\boldsymbol{\theta}) \right)\\
\label{eq:lik_EM_comlete}
&= \sum_{l=1}^{n} \mbox{ln} \left(\sum^{k}_{q=1} \mathds{1}(z_{l}=q) \tau_{q} \mathcal{L}(\lambda_{q}(t)|\boldsymbol{\chi}_{l}) \right),
\end{aligned}
\end{eqnarray}

\noindent and corresponding incomplete data log likelihood is

\begin{eqnarray}
\begin{aligned}
l(\boldsymbol{\theta}|\boldsymbol{D}) &= \sum_{l=1}^{n} \mbox{ln} \left( p(\boldsymbol{\chi}_{l} |\boldsymbol{\theta}) \right) \\
\label{eq:lik_EM_incomlete}
&= \sum_{l=1}^{n} \mbox{ln} \left(\sum^{k}_{q=1} \tau_{q} \mathcal{L}(\lambda_{q}(t)|\boldsymbol{\chi}_{l}) \right).
\end{aligned}
\end{eqnarray}
Here $\tau_{\nu}$ is the `mixing weight' of component $\nu$, where  $0 \leq \tau_{\nu} \leq 1$, $\sum^{k}_{q=1} \tau_{q} =1$, and $\mathcal{L}(\lambda_{\nu}(t)|\boldsymbol{\chi}_{l})$ is the likelihood for the rate function associated with NHPP model $\nu$ given the observation $\boldsymbol{\chi}_{l}$. As before, we represent the rate functions associated with each component $\nu$ as a linear combination of basis functions, i.e. $\lambda_{\nu}(t)=\sum _{m=1}^{n_{b}}c_{\nu,m}B_{m}(t)$.

We use the EM algorithm (see \citet{dempster1977maximum} for details) to find estimates for the parameters $\boldsymbol{\theta}=\{\boldsymbol{\tau}, \boldsymbol{C} \}$ (where $\boldsymbol{\tau}=\{\tau_{1}, \tau_{2}, \ldots , \tau_{k} \}$, $\boldsymbol{C}=\{\boldsymbol{c}_{1}, \boldsymbol{c}_{2}, \ldots, \boldsymbol{c}_{k}\}$ and  $\boldsymbol{c}_{\nu} = \{c_{\nu, 1}, \ldots  , c_{\nu, n_{b}}\}$ is the set of basis function coefficients for NHPP rate function for group $\nu$) which maximise the log likelihood of the mixture. This procedure is described below.\\

\noindent \textbf{Expectation step}\\
Given a current estimate for the model parameters $\widehat{\boldsymbol{\theta}}^{c}$ we find the so called `auxiliary function' which is the expected value of the data log likelihood (\ref{eq:lik_EM_comlete}) of the mixture model with respect to the conditional distribution of $\boldsymbol{z}$ given the data $\boldsymbol{D}$.

\begin{align}
Q(\boldsymbol{\theta}, \widehat{\boldsymbol{\theta}}^{c}) &= \mbox{E}_{\boldsymbol{z} | \boldsymbol{D}, \widehat{\boldsymbol{\theta}}^{c}}[l(\boldsymbol{\theta}|\boldsymbol{D}, \boldsymbol{z})] \nonumber \\ \nonumber
&=\sum_{l=1}^{n} \mbox{E}_{\boldsymbol{z} | \boldsymbol{D}, \widehat{\boldsymbol{\theta}}^{c}}[l(\boldsymbol{\theta}|\boldsymbol{\chi}_{l}, z_{l})],\\ \nonumber
&= \sum_{l=1}^{n}\sum^{k}_{q=1} p(z_{l}=q|\boldsymbol{x_{l}}, \widehat{\boldsymbol{\theta}}^{c}) l(\boldsymbol{\theta} | \boldsymbol{x_{l}}, z_{l}),\\ 
\label{eq:q_theta}
 &=\sum_{l=1}^{n}\sum^{k}_{q=1}\widehat{R}_{q, l}^{c}\left(\mbox{ln}(\tau_{q})+ l(\lambda_{q}(t)|\boldsymbol{\chi}_{l}))\right).
\end{align}

Here $\widehat{R}_{\nu, l}^{c} = p(z_{l}=\nu|\boldsymbol{\chi_{l}}, \widehat{\boldsymbol{\theta}}^{c})$ is the `membership probability' of observation $l$ to group $\nu$. $l(\lambda_{\nu}(t)|\boldsymbol{\chi}_{l}) = - M_{\nu}(t) + \sum_{i=1}^{m_{l}}\lambda_{\nu}(\chi_{l,i})$ is the log likelihood for rate function $\nu$ for observation $l$. Plugging in this latter expression for the log likelihood into (\ref{eq:q_theta}) we get

\begin{eqnarray}
Q(\boldsymbol{\theta}, \widehat{\boldsymbol{\theta}}^{c})=\sum_{l=1}^{n}\sum^{k}_{q=1}\widehat{R}_{q,l}^{c}\left(\mbox{ln}(\tau_{q})-M_{q}(t)+ \sum_{i=1}^{ m_{l}}\mbox{ln}(\lambda_{q}(\chi_{l,i}))\right),\label{eq:q_theta_exp}
\end{eqnarray}


To obtain the membership probabilities for each observation $l$ of model $\nu$ we regard the mixing weights $\boldsymbol{\tau}$ as prior probabilities of each mixture component and use Baye's theorem to obtain

\begin{eqnarray}
\label{eq:EM_mem_prob}
\begin{aligned}
\widehat{R}_{\nu, l}^{c}&=p(z_{l}=\nu|\boldsymbol{\chi}_{l}, \widehat{\boldsymbol{\theta}}^{c})\\
&=\frac{p(z_{l}=\nu )p(\boldsymbol{\chi}_{l} | z_{l}=\nu, \widehat{\boldsymbol{\theta}}^{c})}{\sum_{q=1}^{k} p(z_{l}=q )p(\boldsymbol{\chi}_{l} | z_{l}=q, \widehat{\boldsymbol{\theta}}^{c})}\\
&= \frac{\widehat{\tau}_{\nu}^{c} e^{-\widehat{M}_{\nu}^{c}(t)}\prod_{i=1}^{m_{l}}\widehat{\lambda}_{\nu}^{c}(\chi_{l,i})}{\sum_{q=1}^{k}\widehat{\tau}_{q}^{c} e^{-\widehat{M}^{c}_{q}(t)} \prod_{i=1}^{m_{l}}\widehat{\lambda}^{c}_{q}(\chi_{l,i})}.
\end{aligned}
\end{eqnarray}

\noindent \textbf{Maximisation step}\\
After obtaining values for the membership probabilities, we obtain subsequent estimates for the parameters $\widehat{\boldsymbol{\theta}}^{c+1}$ by finding the $\boldsymbol{\theta}$ which maximise the auxiliary function (\ref{eq:q_theta_exp}) subject to the constraint that the mixing weights must sum to 1, i.e. $\sum^{k}_{q=1}\tau_{q}=1$. To address this constraint we consider the Lagrange function

\begin{eqnarray}
\label{eq:lagrange}
\mathcal{L}(\boldsymbol{\theta}, \widehat{\boldsymbol{\theta}}^{c},\delta)= \sum_{l=1}^{n}\sum^{k}_{q=1}\widehat{R}_{q,l}^{c}\left(\mbox{ln}(\tau_{q})-M_{q}(t)+ \sum_{i=1}^{ m_{l}}\mbox{ln}(\lambda_{q}(\chi_{l,i}))\right) + \delta\left(\sum^{k}_{q=1}\tau_{q}-1\right),
\end{eqnarray}

\noindent where $\delta$ is the Lagrange multiplier. Updates for the mixing weights $\boldsymbol{\tau}$ are found by taking the partial derivative of (\ref{eq:lagrange}) with respect to $\tau_{\nu}$, setting it to zero and solving for $\boldsymbol{\tau}$. The partial derivative of $\mathcal{L}(\boldsymbol{\theta}, \boldsymbol{\theta}^{c}, \delta)$ is

\begin{align}
\frac{\partial \mathcal{L}(\boldsymbol{\theta}, \boldsymbol{\theta}^{c},\delta)}{\partial \tau_{\nu}}=&
\frac{\partial}{\partial \tau_{\nu}}\sum_{l=1}^{n}\sum^{k}_{q=1}\widehat{R}_{q,l}^c\left(\mbox{ln}(\tau_{q})-M_{q}(t) + \sum_{i=1}^{ m_{l}}\mbox{ln}(\lambda_{q}(\chi_{l,i}))\right)\nonumber \\ 
&+ \frac{\partial}{\partial \tau_{\nu}}\delta\left(\sum^{k}_{q=1}\tau_{q}-1\right). \label{eq:EM_max_der_wrt_tau}
\end{align}

Evaluating (\ref{eq:EM_max_der_wrt_tau}) and removing all constant terms we get

\begin{align}
\frac{\partial \mathcal{L}(\boldsymbol{\theta}, \boldsymbol{\theta}^{c}, \delta)}{\partial \tau_{\nu}}&=\sum_{l=1}^{n}\frac{d}{d \tau_{\nu}}\widehat{R}_{\nu,l}^c\mbox{ln}(\tau_{\nu})+ \frac{d}{d\tau_{\nu}}\delta\tau_{\nu}, \nonumber \\
&=\sum_{l=1}^{n}\frac{\widehat{R}_{\nu,l}^c}{\tau_{\nu}}+ \delta. \label{eq:dqdt}
\end{align}

Setting equation (\ref{eq:dqdt}) to 0 and solving for $\tau_{\nu}$ we get

\begin{eqnarray}
\tau_{\nu}=\sum_{l=1}^{n}\frac{\widehat{R}_{\nu,l}^c}{-\delta}.
\end{eqnarray}

Using the constraint that $\sum^{k}_{q=1}\tau_{q}=1$, then $-\delta=\sum_{l=1}^{n}\sum^{k}_{q=1}\widehat{R}_{q,l}^c=\sum_{l=1}^{n} 1=n$. Hence the updated estimate for the mixing weight is $\widehat{\tau}_{\nu}^{c+1}=\sum_{l=1}^{n}\frac{\widehat{R}_{\nu,l}^c}{n}$.

Next we seek the $\boldsymbol{C}$ which maximise $\mathcal{L}(\boldsymbol{\theta}, \boldsymbol{\theta}^{c}, \delta)$ subject to the constraints that all rate functions are non-negative, i.e.

\begin{eqnarray}
\label{eq:const_EM}
\sum _{m=1}^{n_{b}}c_{q,m}B_{m}(t), \ldots, \sum _{m=1}^{n_{b}}c_{k,m}B_{m}(t) \geq 0,
\end{eqnarray}

\noindent for all $q=1, \ldots, k$.

This is a convex minimisation problem of the same variety as encountered in Section \ref{sec:rate_func}. To see this, consider the term $- f(\boldsymbol{c}_{\nu, l})$= \\
\noindent $-M_{q}(t) + \sum_{i=1}^{ m_{l}}\mbox{ln}\left(\sum _{m=1}^{n_{b}}c_{q,m}B_{m}(\chi_{l,i})\right)$ from equation (\ref{eq:lagrange}). $f(\boldsymbol{c}_{\nu}, l)$ is identical to equation (\ref{eq:log_lik_max}) which we showed is convex in the set of basis function coefficients in Section \ref{sec:rate_func}. As $\widehat{R}_{\nu, l} \geq 0$ for every $\nu$ and $l$, $\widehat{R}_{\nu,l}f(\boldsymbol{c}_{\nu, l}) = $\\
\noindent$\widehat{R}_{\nu,l}\left(M_{q}(t) - \sum_{i=1}^{ m_{l}}\mbox{ln}(\sum _{m=1}^{n_{b}}c_{q,m}B_{m}(\chi_{l,i}))\right)$ is also convex. Furthermore, disregarding terms which do not involve $\boldsymbol{C}, -\mathcal{L}(\boldsymbol{\theta},\widehat{\boldsymbol{\theta}}^{c}, \delta)=\sum_{l=1}^{n}\sum^{k}_{q=1}\widehat{R}_{q,l}^{c}f(\boldsymbol{c}_{\nu, l})$ is convex in $\boldsymbol{C}$ as sums of convex functions are also convex. The functions in the constraints given in (\ref{eq:const_EM}) are linear in $\boldsymbol{C}$ and hence convex. Therefore to find the $\boldsymbol{C}$ which maximise $\mathcal{L}(\boldsymbol{\theta}, \boldsymbol{\theta}^{c},\delta)$ (\ref{eq:lagrange}) we can use the same numerical approach as used to minimise (\ref{eq:log_lik_max}). The expectation and maximisation steps are applied iteratively until the the change in value of the Lagrangian function from one step to the next is less than a prescribed threshold. We used a value of 1e-4 for the results in this paper.  \\

\noindent \textbf{Initialisation}\\
\noindent Depending on the starting conditions the EM algorithm may converge to a local optimum. To overcome this issue we use a form of the random restart approach commonly used with the EM algorithm for Gaussian mixtures \citep{biernacki2003choosing} described as follows. To set initial estimates for $\boldsymbol{C}$, each observation $\boldsymbol{\chi}_{l}$ is randomly assigned a class label $g_{l}'\in{1,2,\ldots,k}$. Rate function estimates for the `random' classes are obtained using the classification procedure outlined in Section \ref{sec:classification} and the basis function estimates which correspond to these are used for the initial estimates for $\boldsymbol{C}$. All mixing weights in $\boldsymbol{\tau}$ are initially set to 1/$k$. The EM algorithm is then initialised from a number of different sets of initial conditions corresponding to different `random' class assignments and the solution with the largest log likelihood is retained. Using this procedure, for all the examples considered in this paper, 3 random restarts were typically plenty to avoid local maxima.

\section{Empirical results} \label{sec:data}
In this section we present the results of our methods applied to various synthetic and real world data sets. 

\subsection{Synthetic data}\label{sec:synthetic_results}
We first applied our methods to synthetically generated event data with prescribed rate functions. None of the data generating rate functions belong to the class of smooth functions we assumed in our methods. We generated three difference data sets based on the following four rate functions.

\begin{align}
\nonumber
\hspace*{-0cm} \lambda_{1}(t)&=100\mbox{sin}^{2}(t/2) \hspace{2cm} \lambda_{2}(t)=100\mbox{sin}^{2}(t)
\end{align}
\begin{multicols}{2}

\[
  \lambda_{3}(t)=\left\{\def\arraystretch{1.2}%
  \begin{array}{@{}c@{\quad}l@{}}
    20 & \text{if $t < \frac{t_{2}}{4}$}\\
    40 & \text{if $\frac{t_{2}}{4} \leq t < \frac{t_{2}}{2}$}\\
    60 & \text{if $\frac{t_{2}}{3} \leq t < \frac{3t_{2}}{4}$}\\
    80 & \text{if $t \geq \frac{3t_{2}}{4}$}\\
  \end{array}\right.
\]

\columnbreak

\[
  \lambda_{4}(t)=\left\{\def\arraystretch{1.2}%
  \begin{array}{@{}c@{\quad}l@{}}
    80 & \text{if $t < \frac{t_{2}}{4}$}\\
    60 & \text{if $\frac{t_{2}}{4} \leq t < \frac{t_{2}}{2}$}\\
    40 & \text{if $\frac{t_{2}}{3} \leq t < \frac{3t_{2}}{4}$}\\
    20 & \text{if $t \geq \frac{3t_{2}}{4}$}\\
  \end{array}\right.
\]

\end{multicols}

For synthetic data set 1 we generated event times for observations with class label 1 (\emph{resp.} 2) separately using rate $\lambda_{1}(t)$ ($\lambda_{2}(t)$). As $\lambda_{1}(t)$ and $\lambda_{2}(t)$ are both periodic functions this experiment mimics real world examples of point processes which exhibit periodic patterns (e.g. time of stock trades \citep{engle2000econometrics} and the times at which emails are sent \citep{malmgren2008poissonian}).

For synthetic data set 2, $\lambda_{3}(t)$ (\emph{resp.} $\lambda_{4}(t)$) was used to generate class 1 (2) observations. These rate functions mimic computer network traffic for which a NHPP model with rate function defined by a step function is a suitable model \citep{paxson1995wide}.

Finally for data set 3 we considered a four class problem where rate functions $\lambda_{1}(t), \lambda_{2}(t), \lambda_{3}(t)$ and $\lambda_{4}(t)$ were used to generate data in classes 1,2,3 and 4 respectively. The thinning algorithm \citep{lewis1978simulation} was used to generate all data. Synthetic event time data for 20 observations for each rate function is shown in Figure \ref{fig:synthetic_plots}.

\subsubsection{Classification}
For each class, event time data for 20 observations was generated. 10 of these formed the training set and the remainder formed the unseen test set. Using the classification procedure outlined in Section \ref{sec:classification} we were able to recover the prescribed rate function for each data set (see Figure \ref{fig:synthetic_plots}). Furthermore, when each observation in the test data was assigned to the class for which its membership probability (as defined in equation (\ref{eq:mp_class})) is maximal, the classification accuracy was 100\%.  
 
\subsubsection{Clustering}
\begin{figure*}[htbp]
\begin{center}
\textbf{Synthetic data event times}\\
\includegraphics[width=0.24\textwidth]{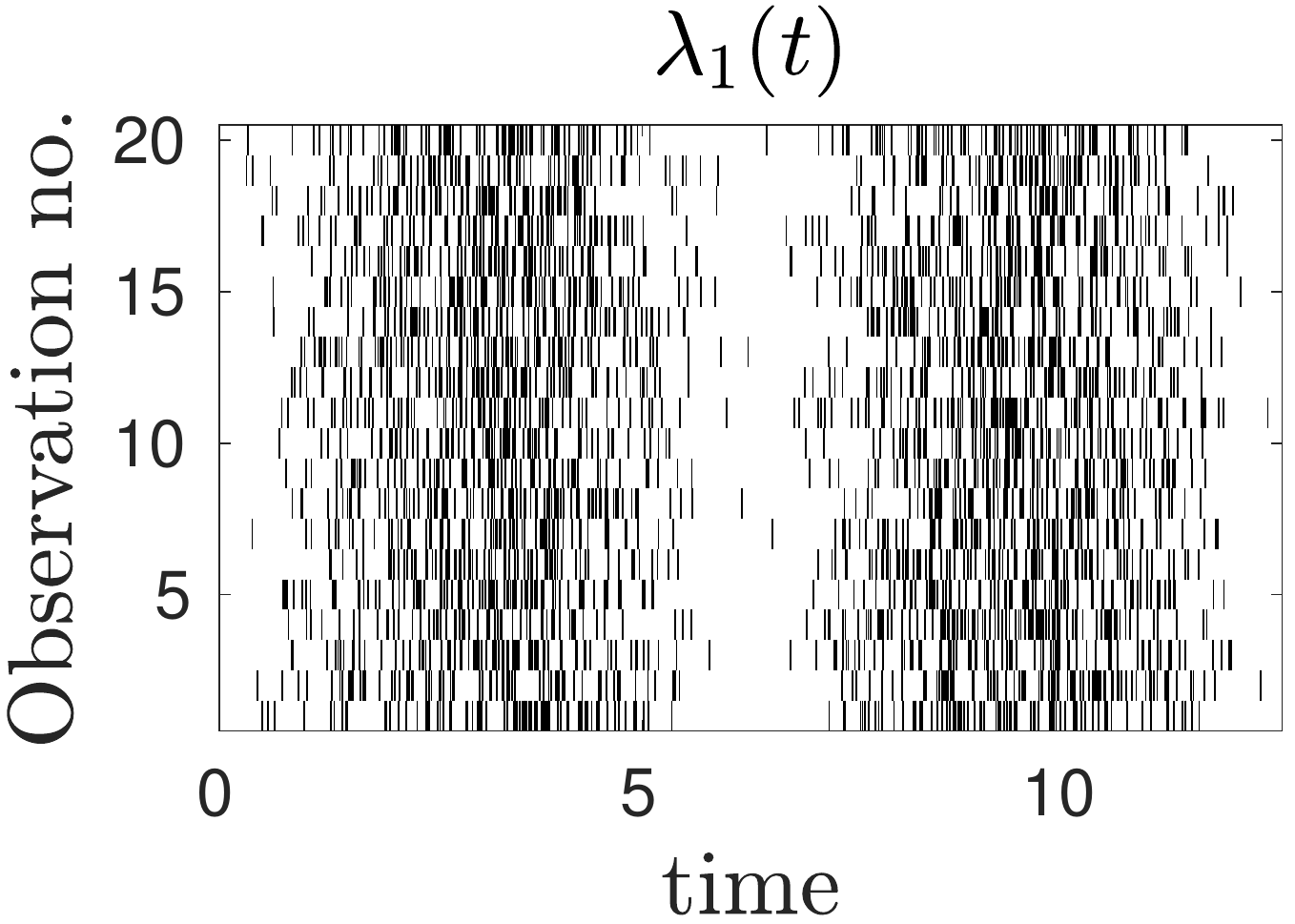}
\includegraphics[width=0.24\textwidth]{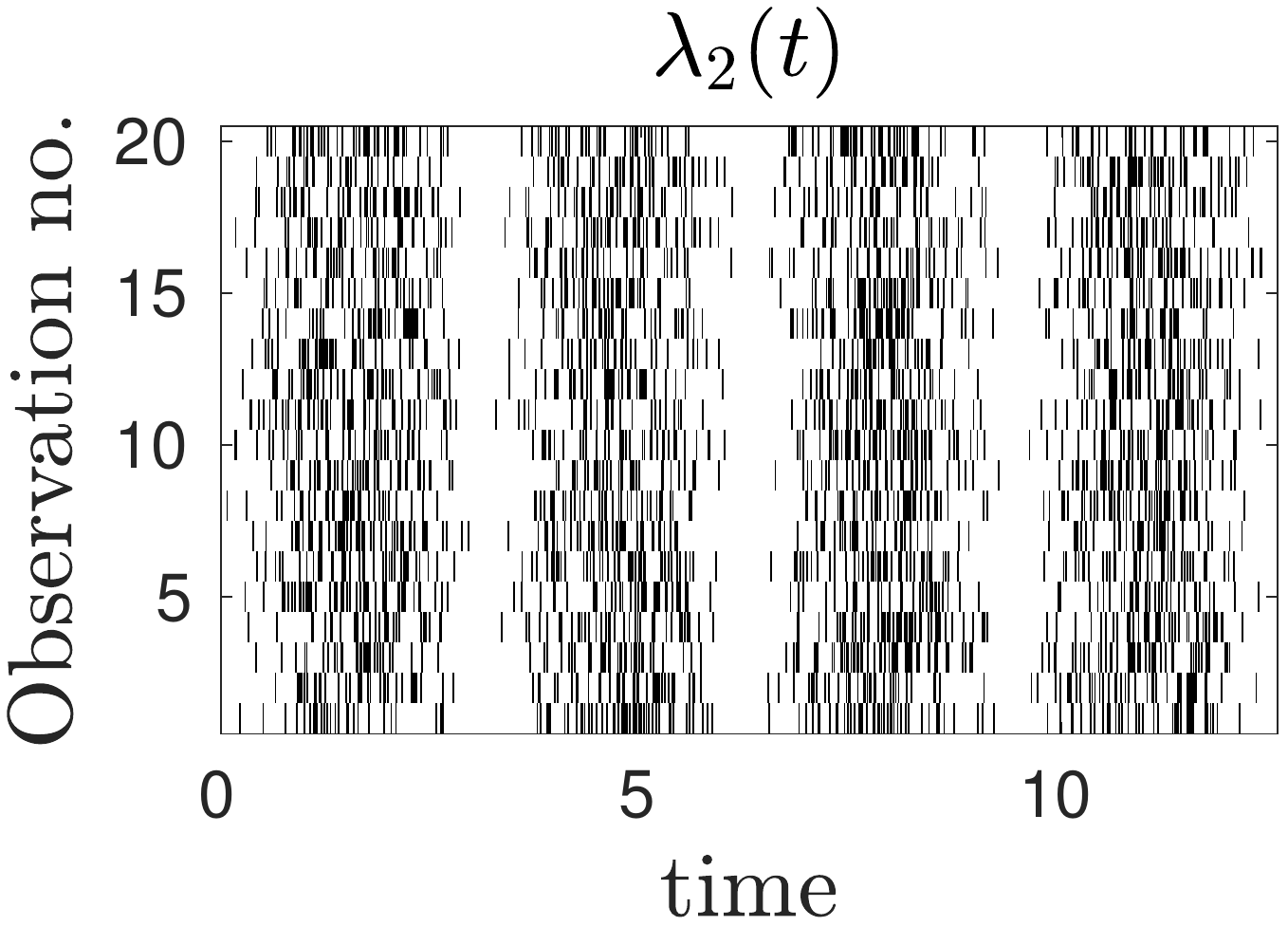}
\includegraphics[width=0.24\textwidth]{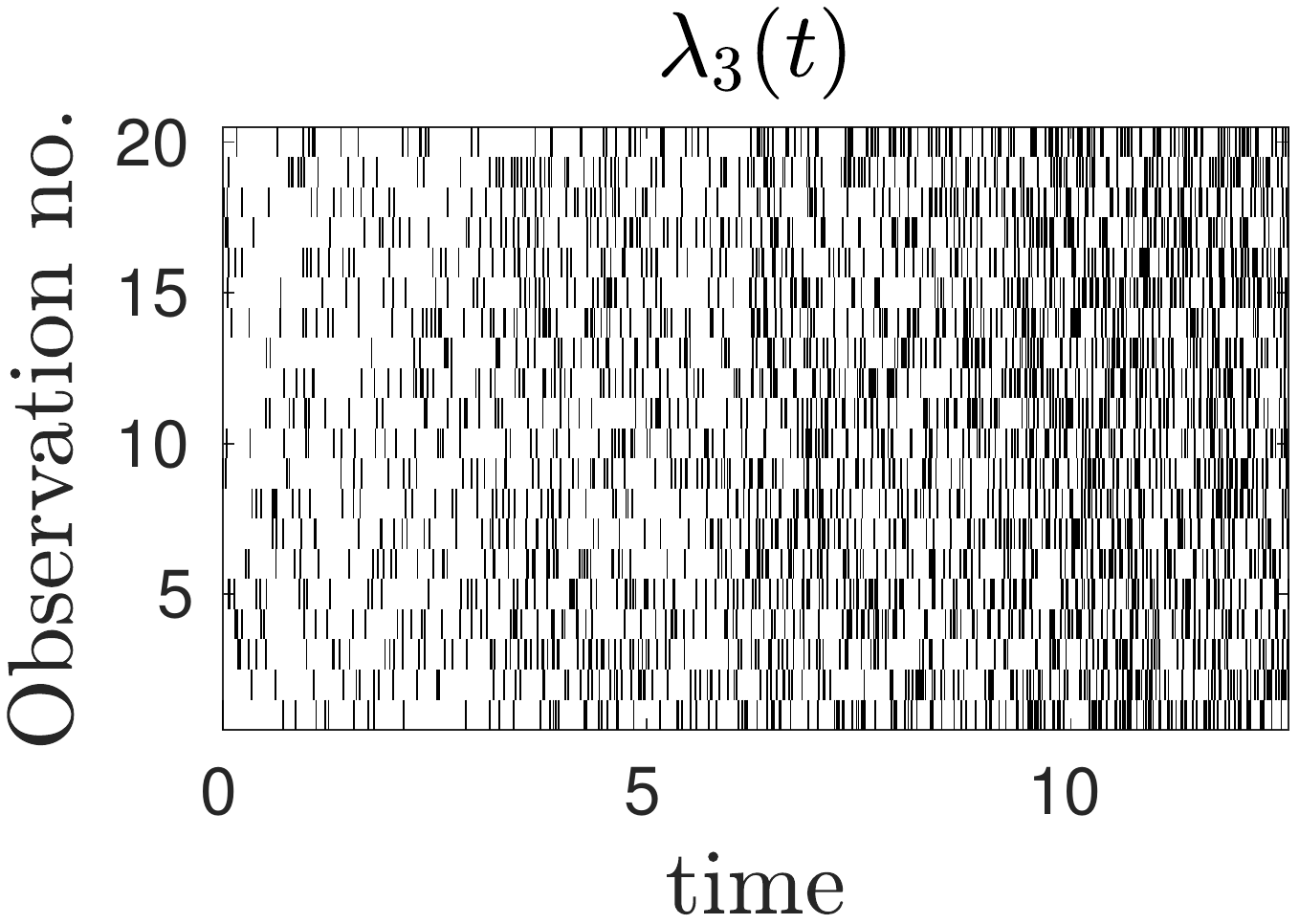}
\includegraphics[width=0.24\textwidth]{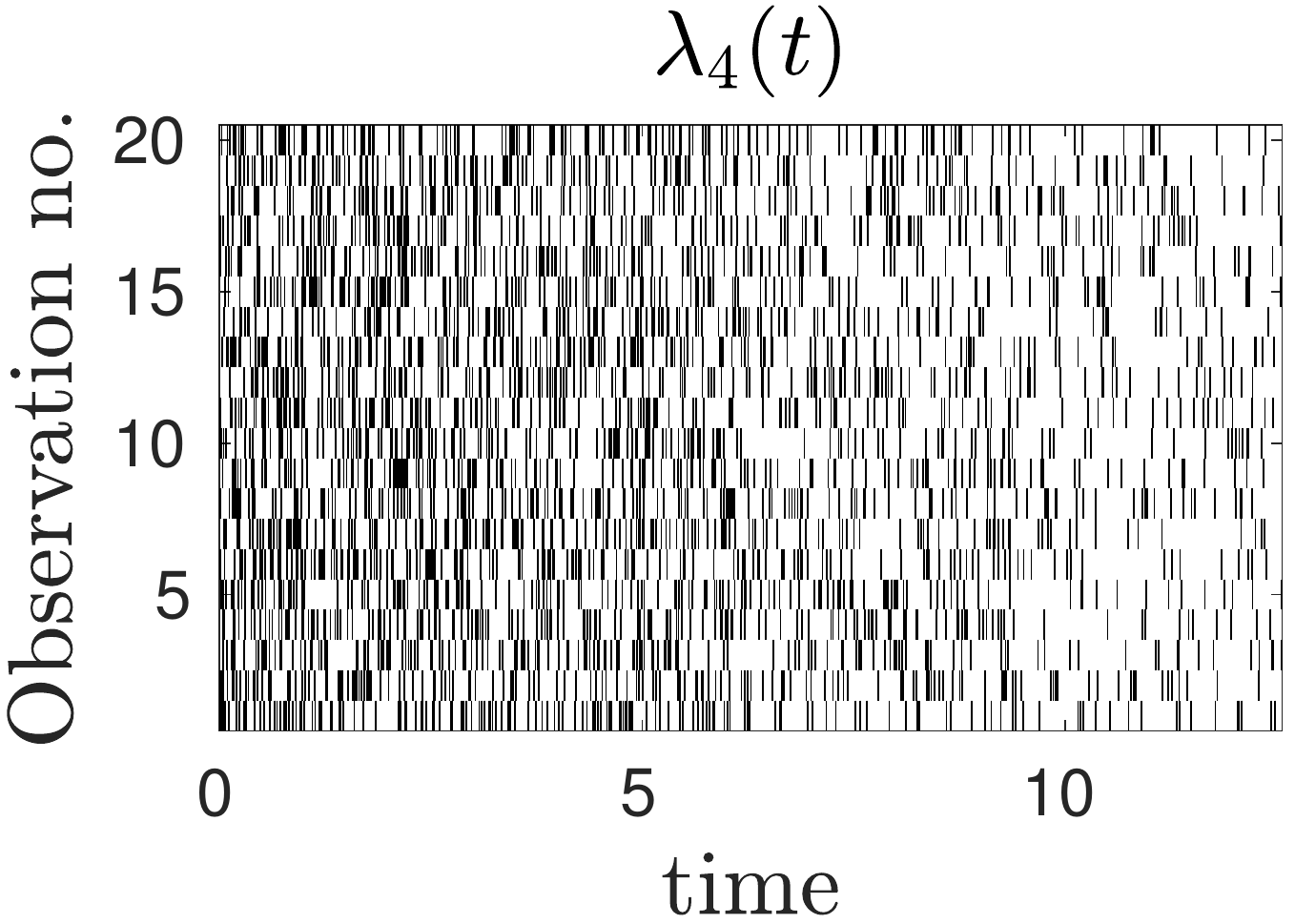}\\
\hspace{0.5cm} \textbf{Data set 1 - classification} \hspace{2cm} \textbf{Data set 2 - classification}\\
\includegraphics[width=0.24\textwidth]{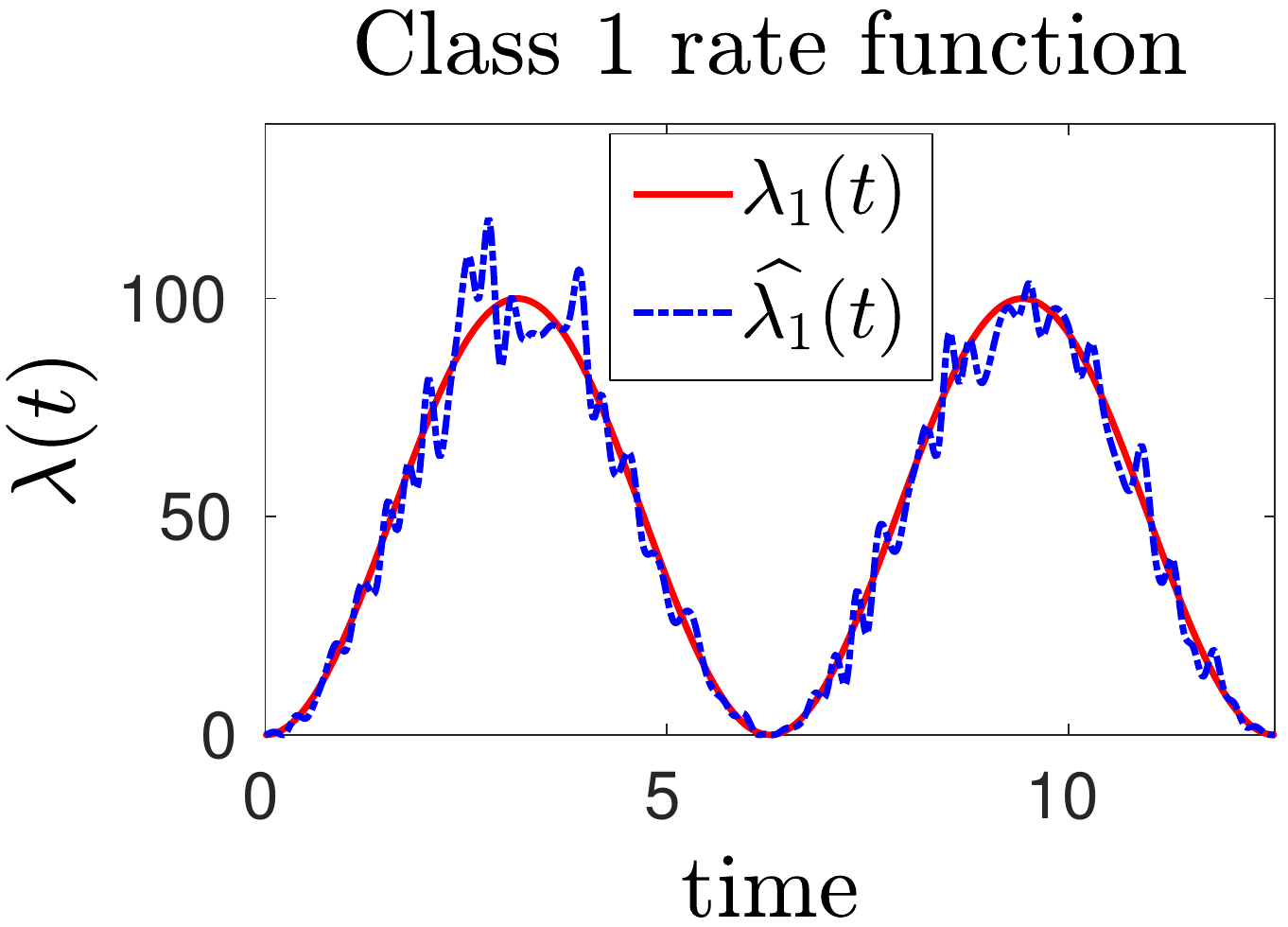}
\includegraphics[width=0.24\textwidth]{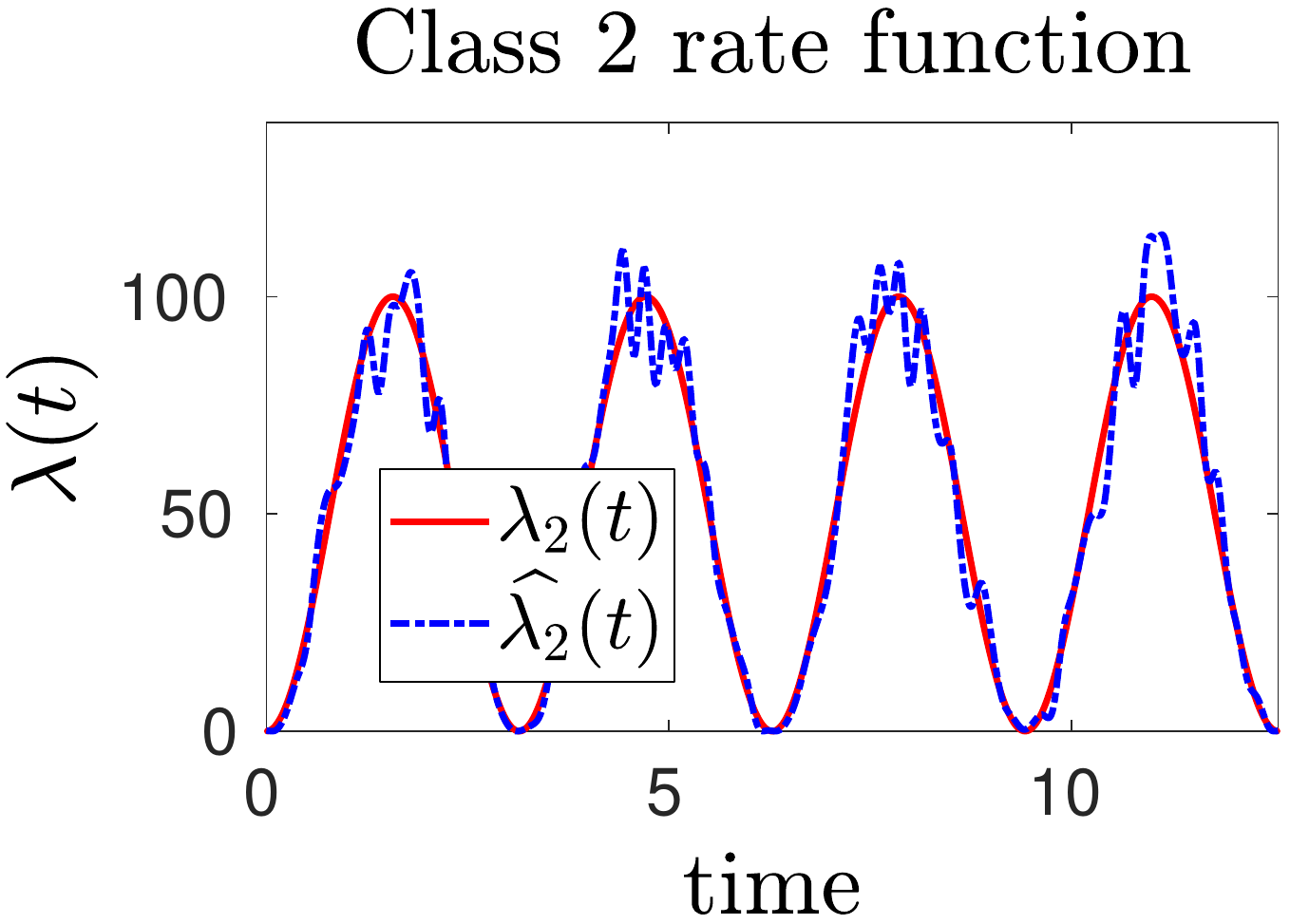}
\includegraphics[width=0.24\textwidth]{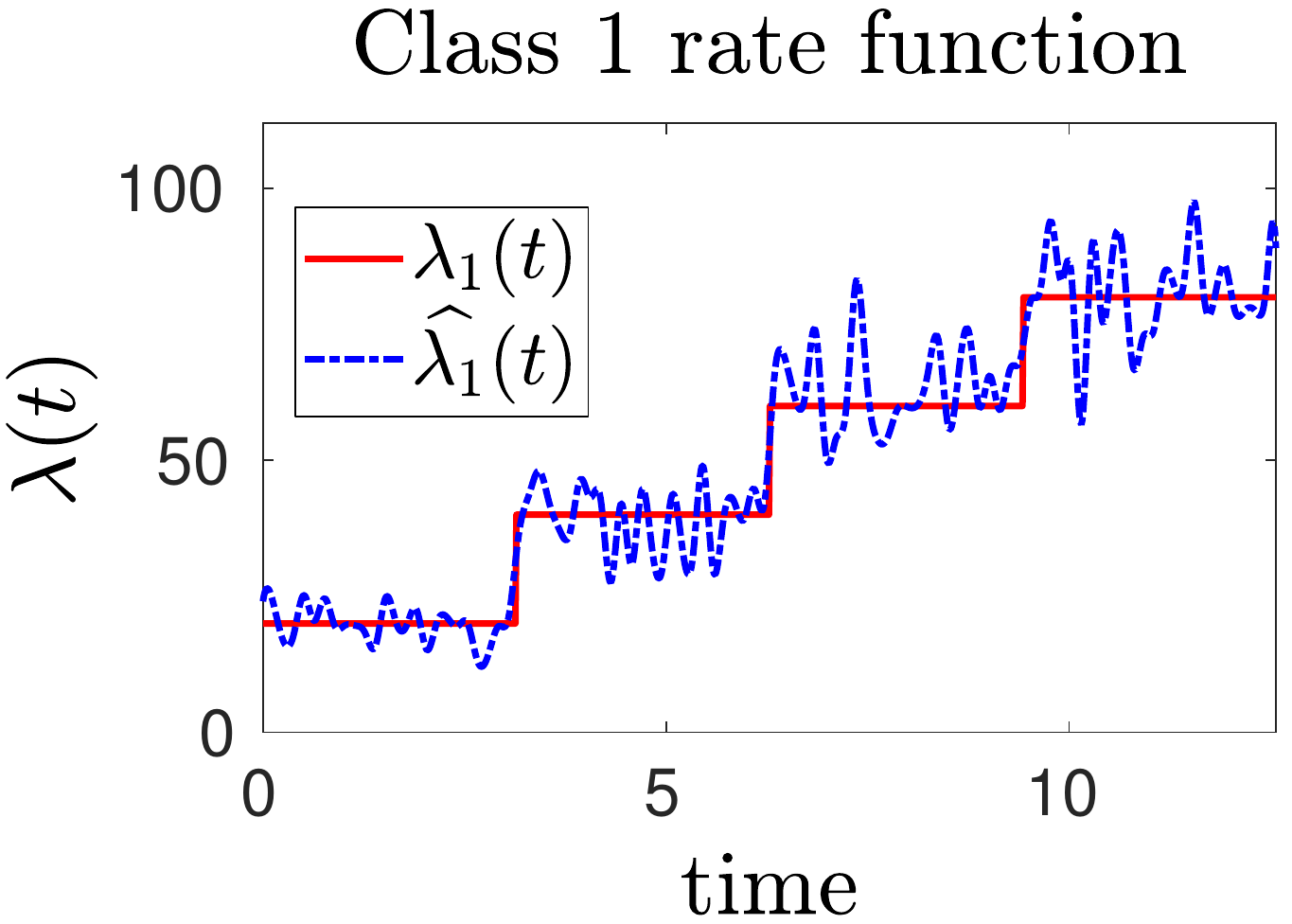}
\includegraphics[width=0.24\textwidth]{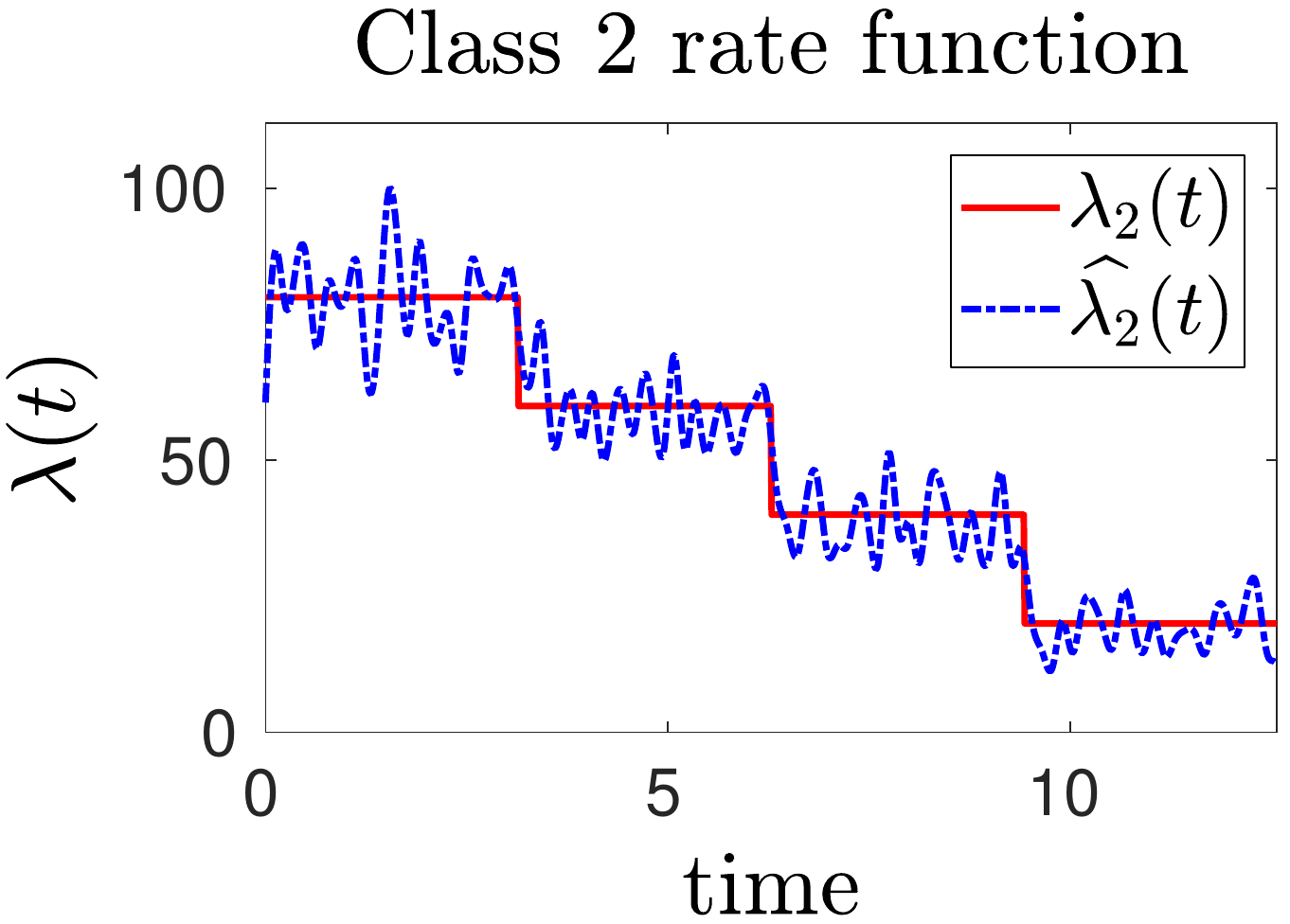}\\
\textbf{Synthetic data set 3 - classification task}\\
\includegraphics[width=0.24\textwidth]{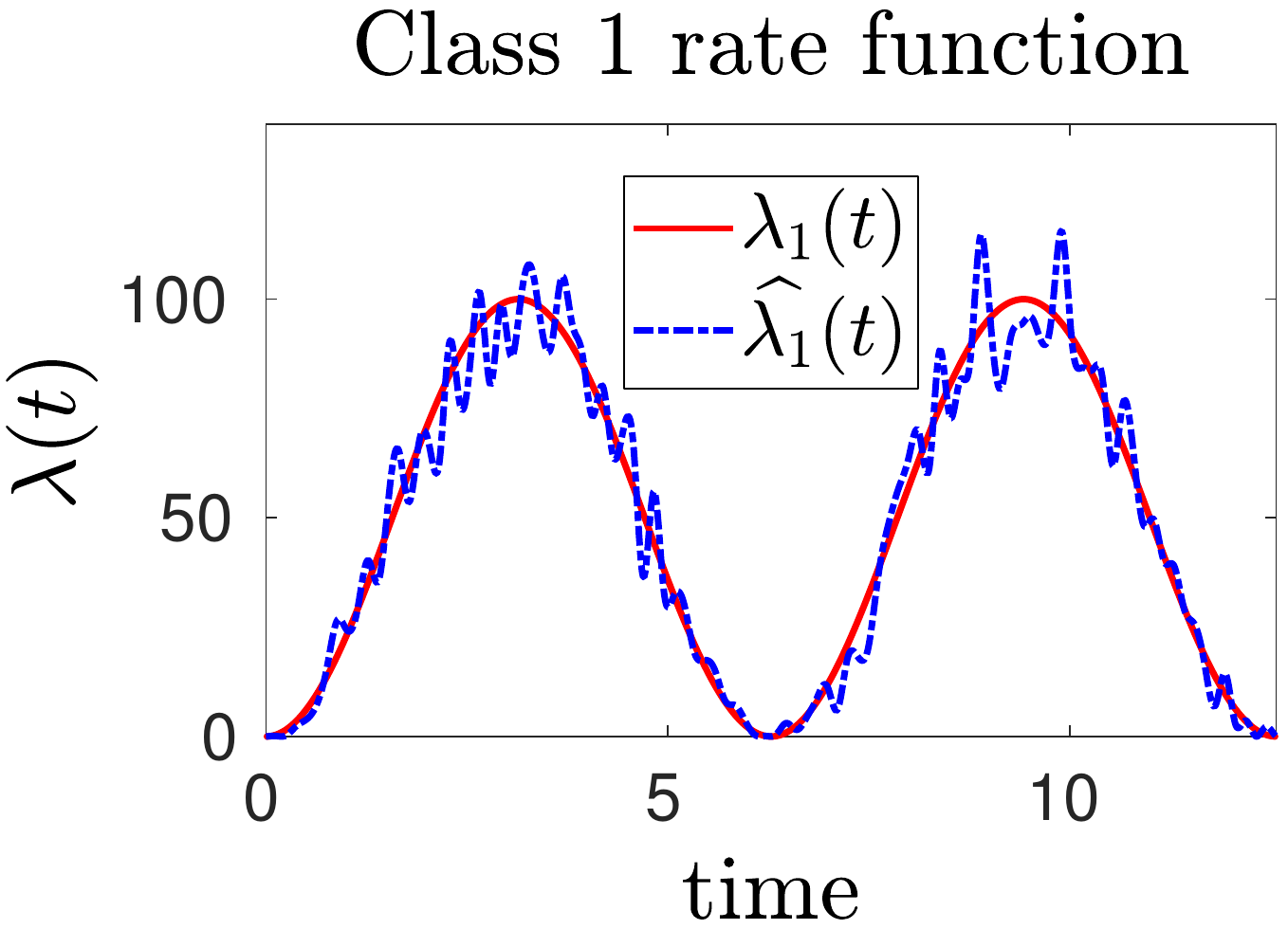}
\includegraphics[width=0.24\textwidth]{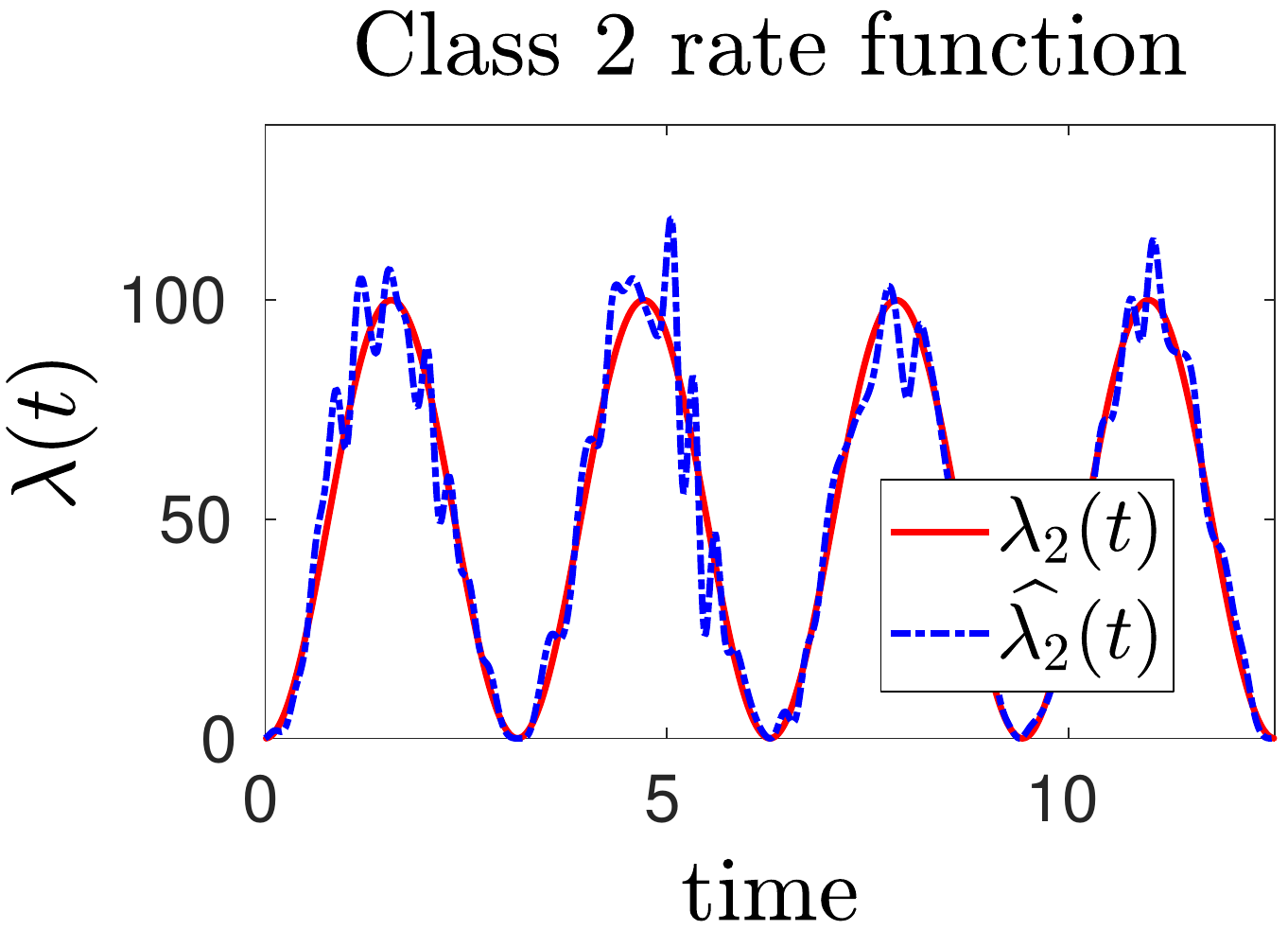}
\includegraphics[width=0.24\textwidth]{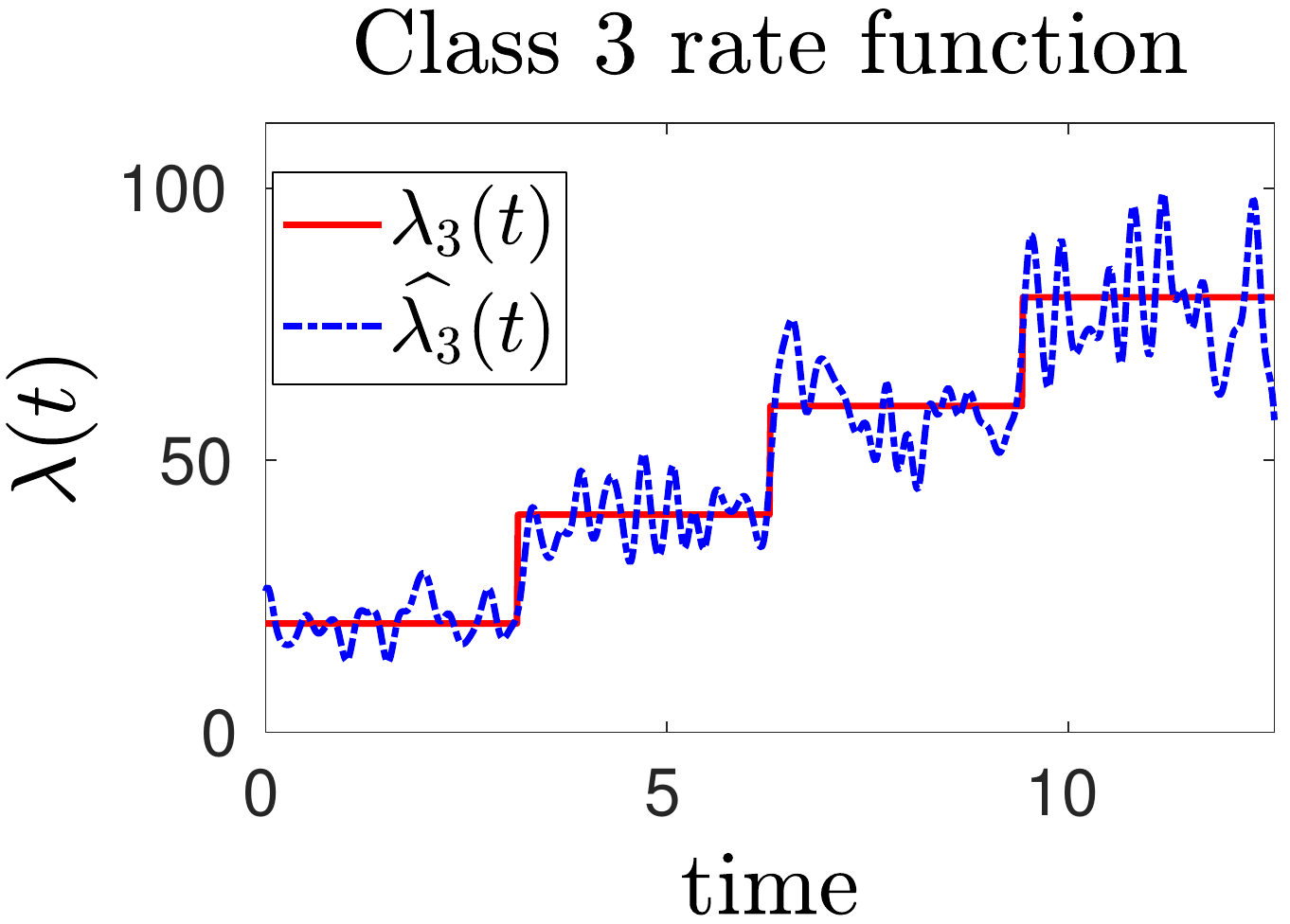}
\includegraphics[width=0.24\textwidth]{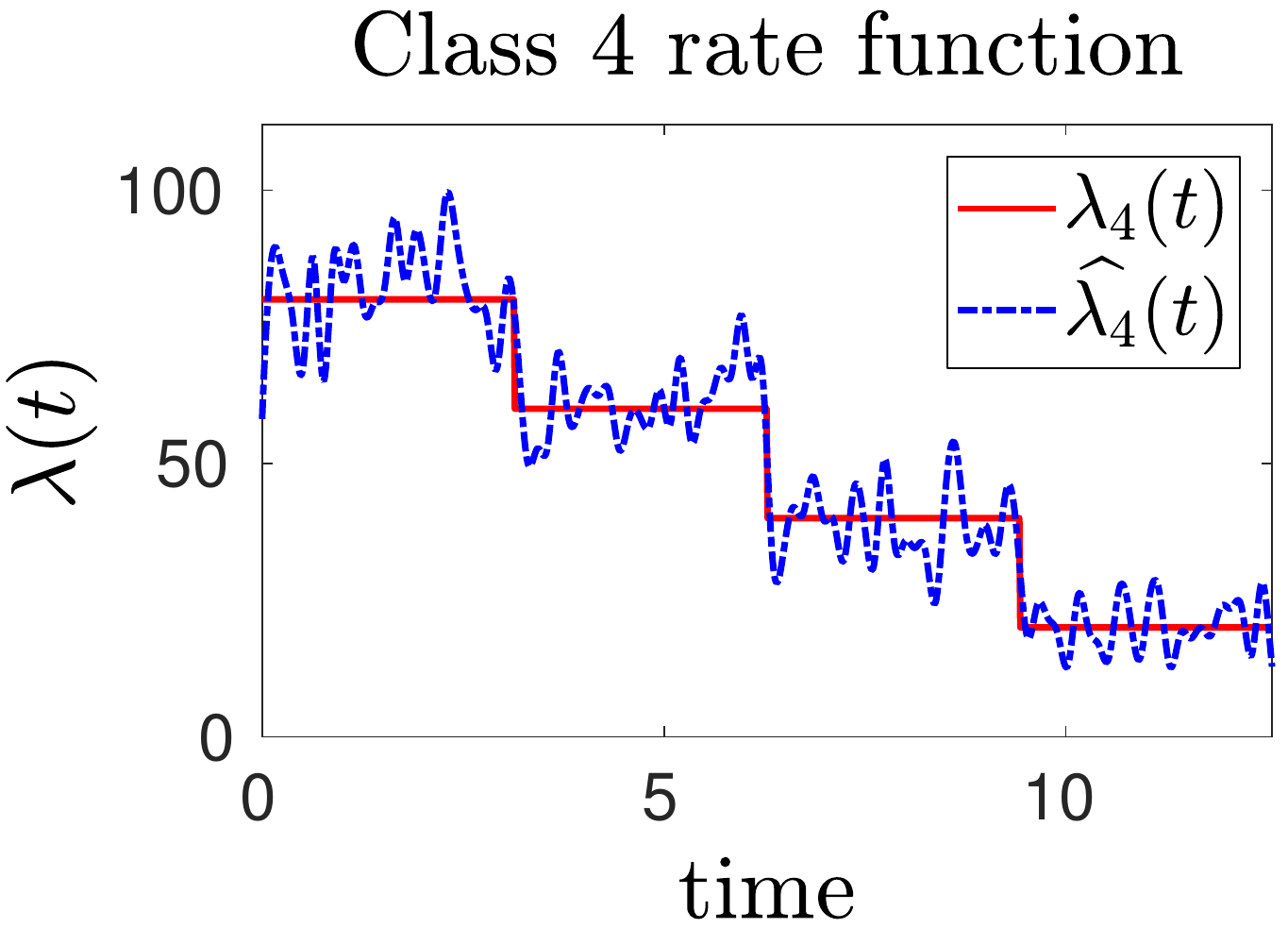}\\
\hspace{0.5cm} \textbf{Data set 1 - clustering} \hspace{2cm} \textbf{Data set 2 - clustering}\\
\includegraphics[width=0.24\textwidth]{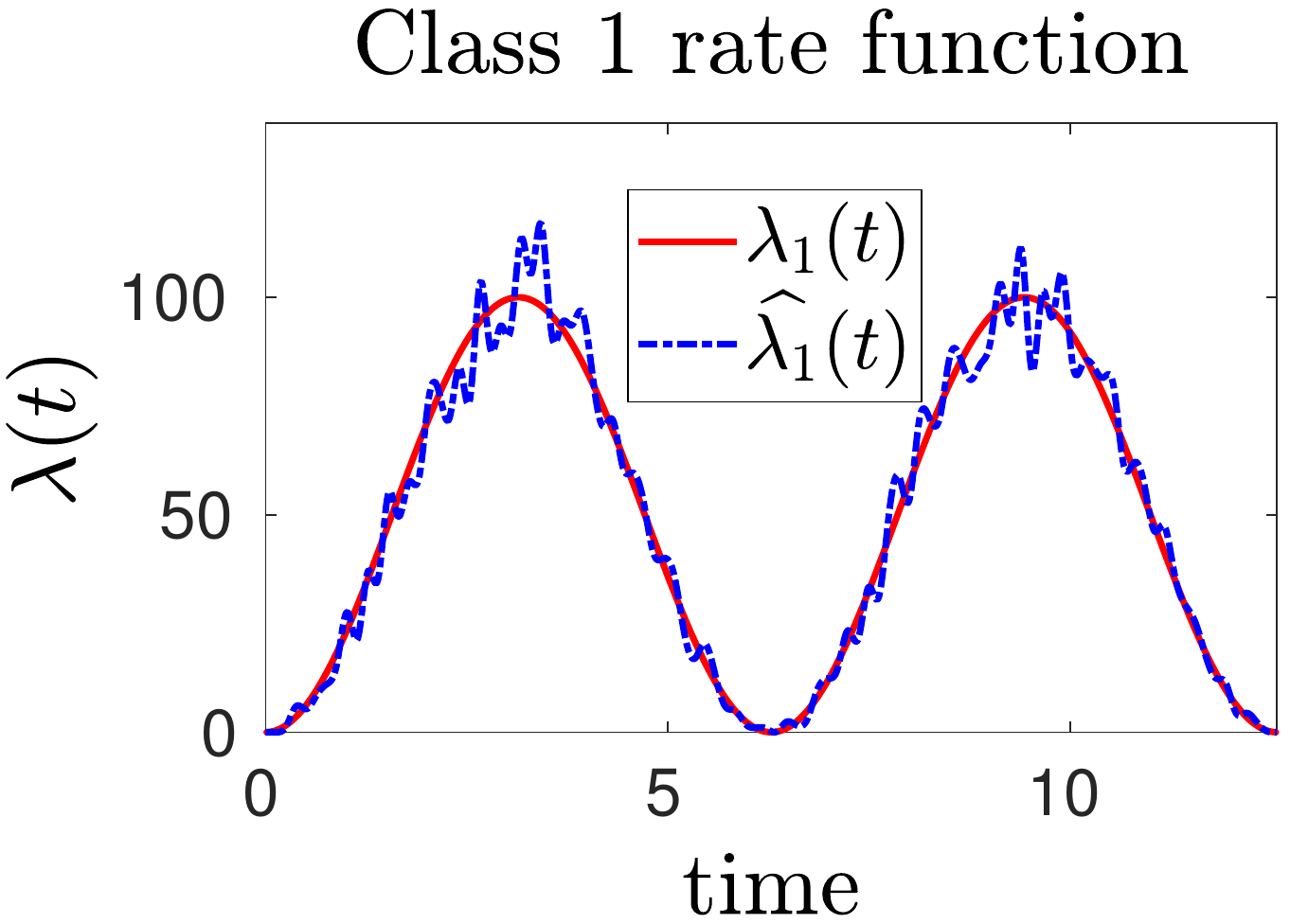}
\includegraphics[width=0.24\textwidth]{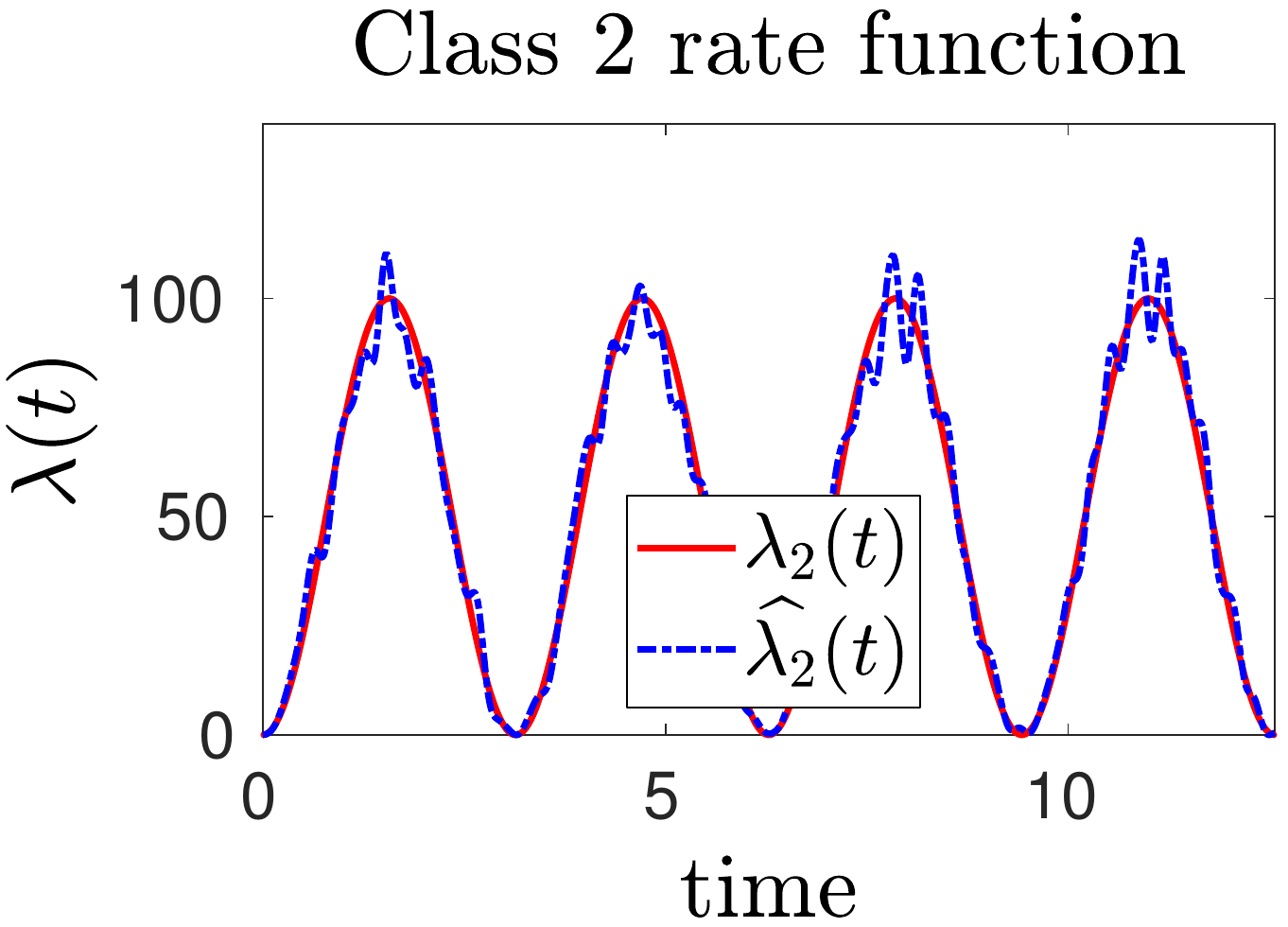}
\includegraphics[width=0.24\textwidth]{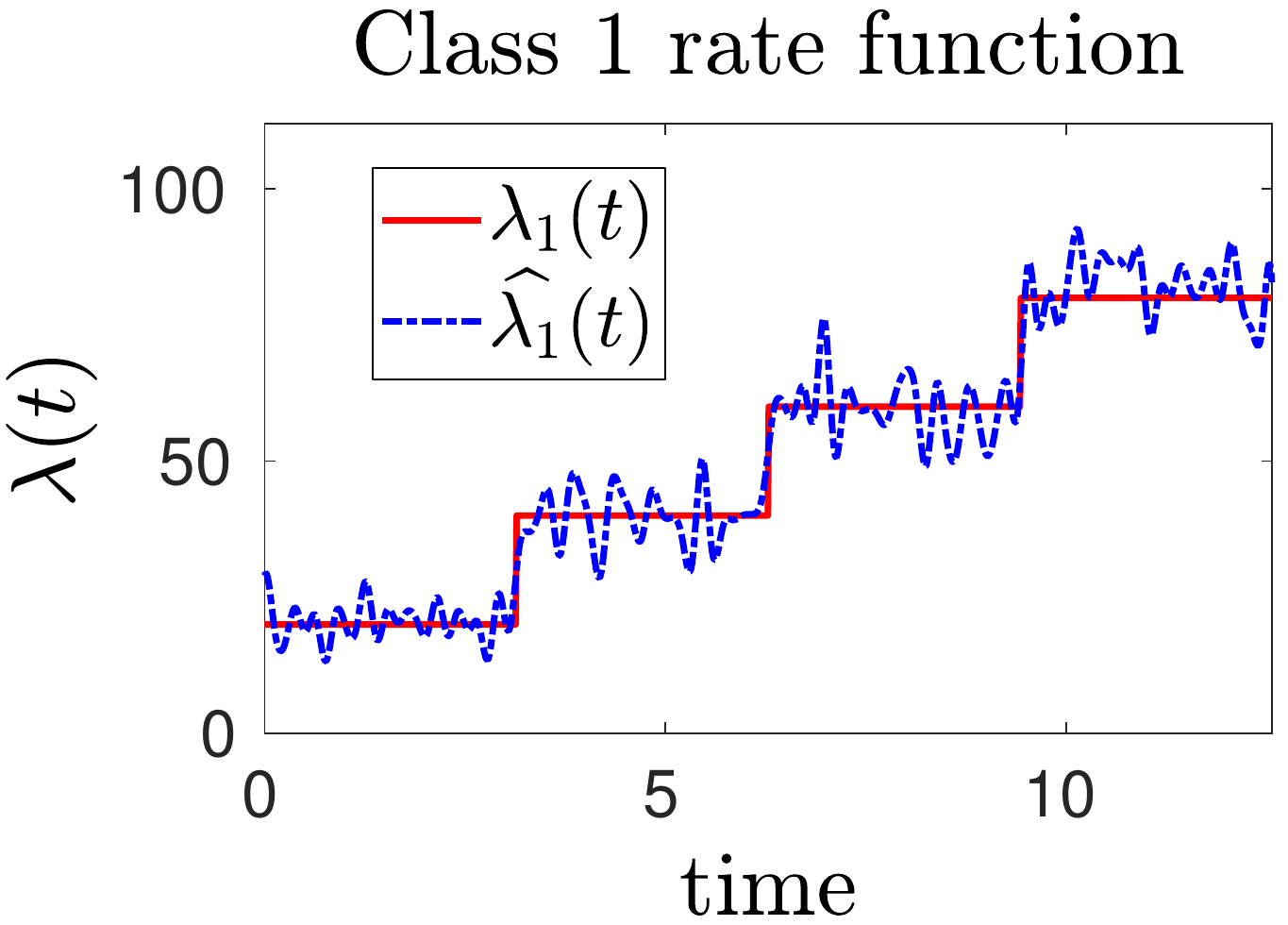}
\includegraphics[width=0.24\textwidth]{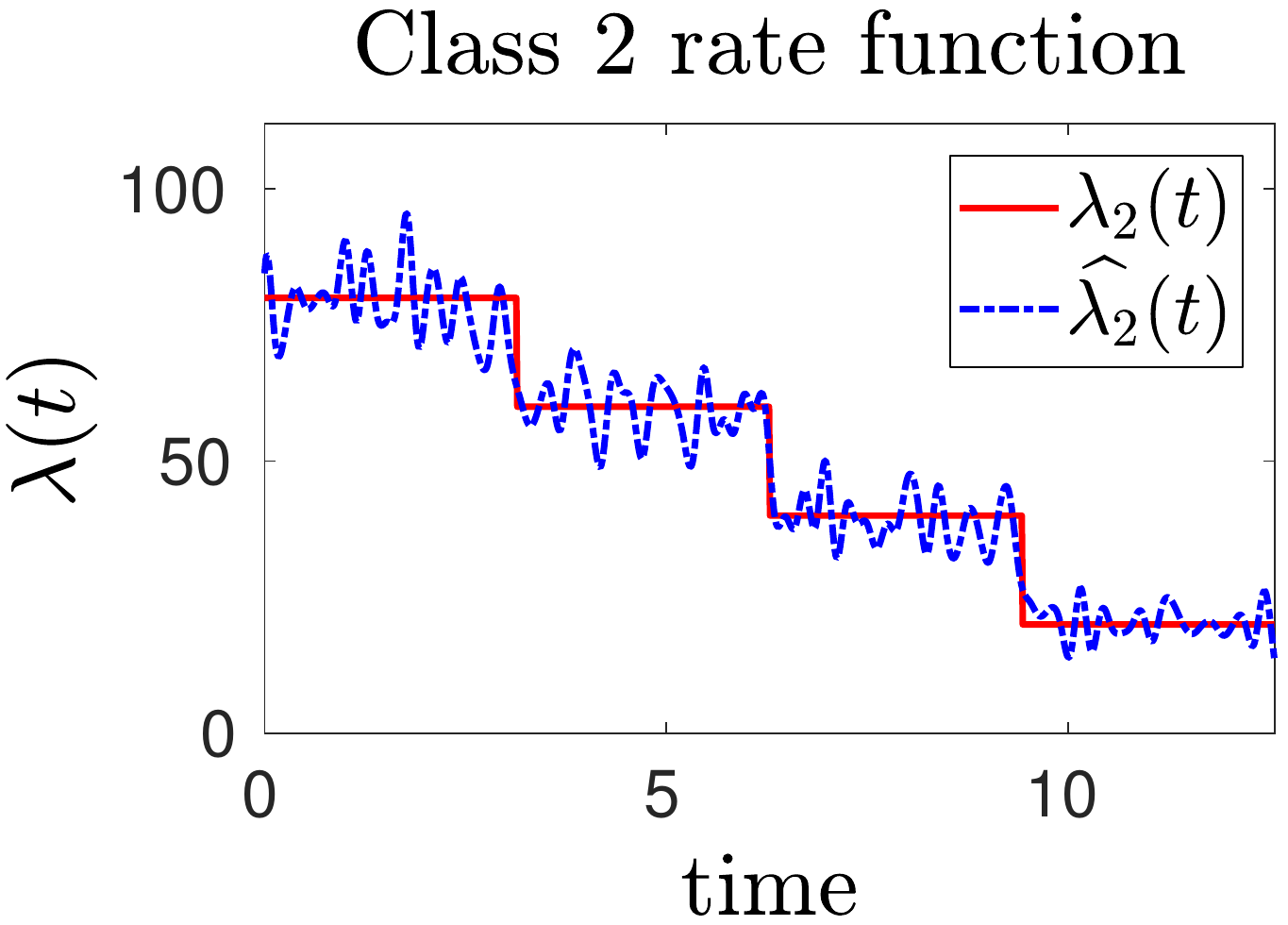}\\
\textbf{Synthetic data set 3 - clustering task}\\
\includegraphics[width=0.24\textwidth]{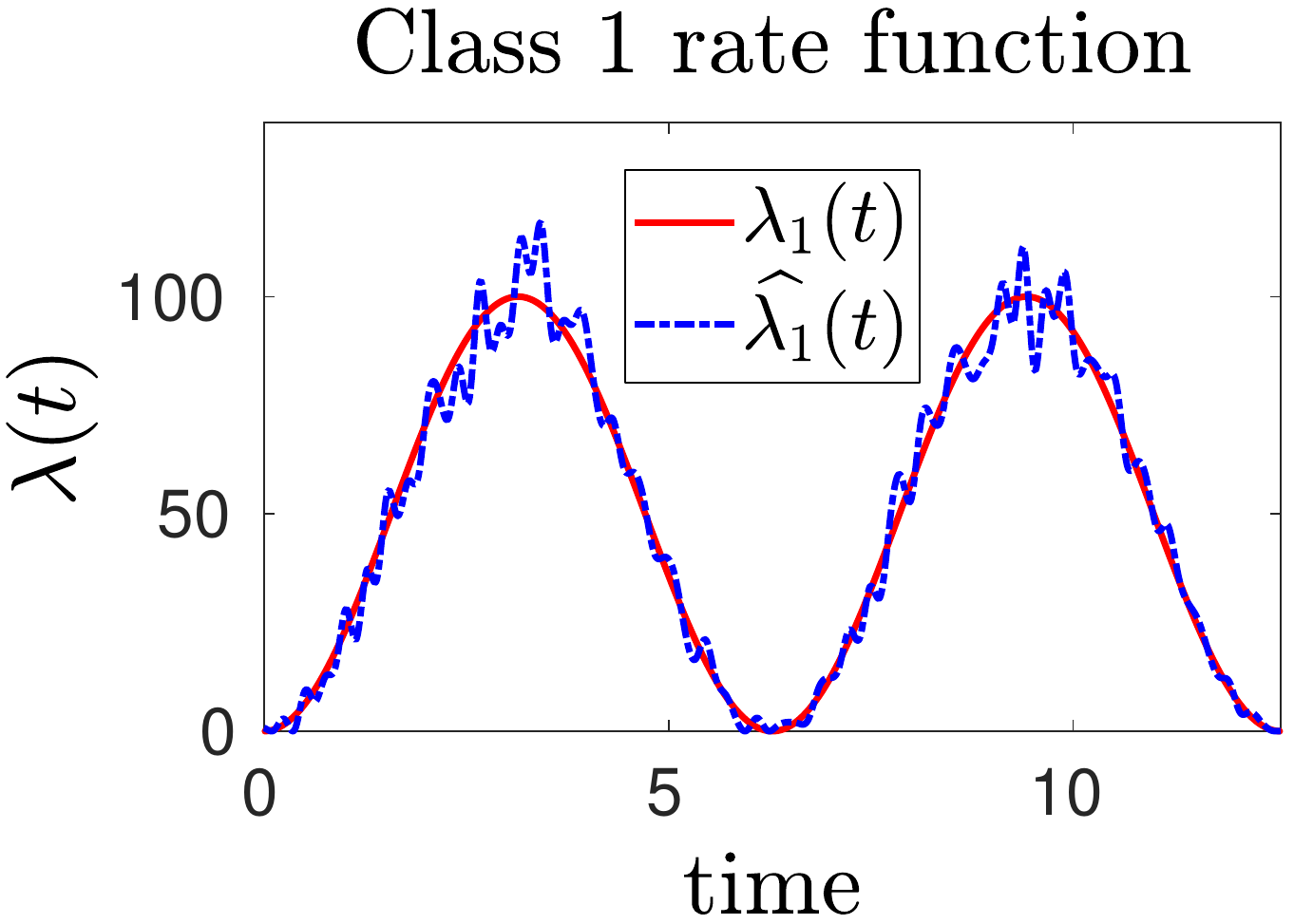}
\includegraphics[width=0.24\textwidth]{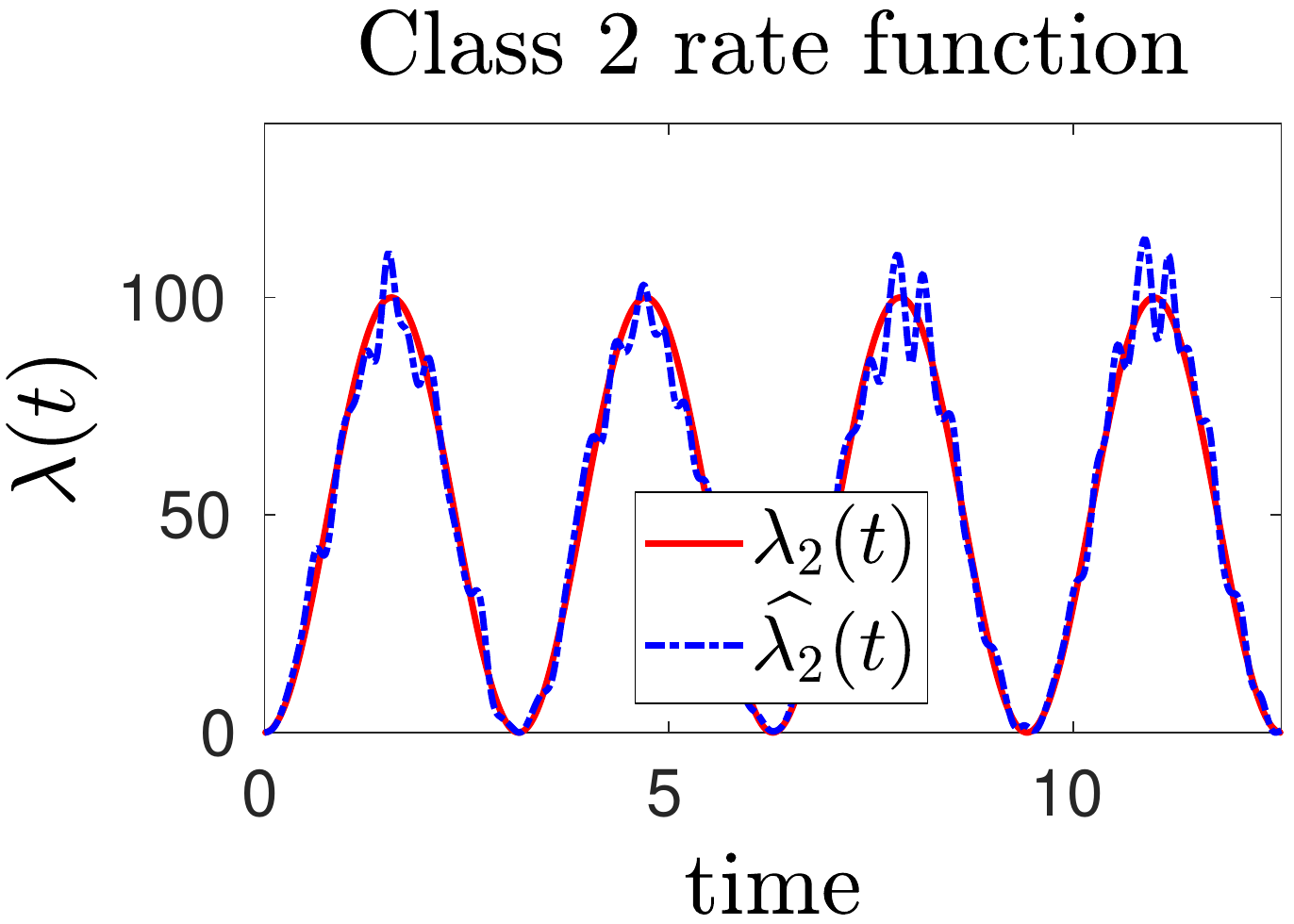}
\includegraphics[width=0.24\textwidth]{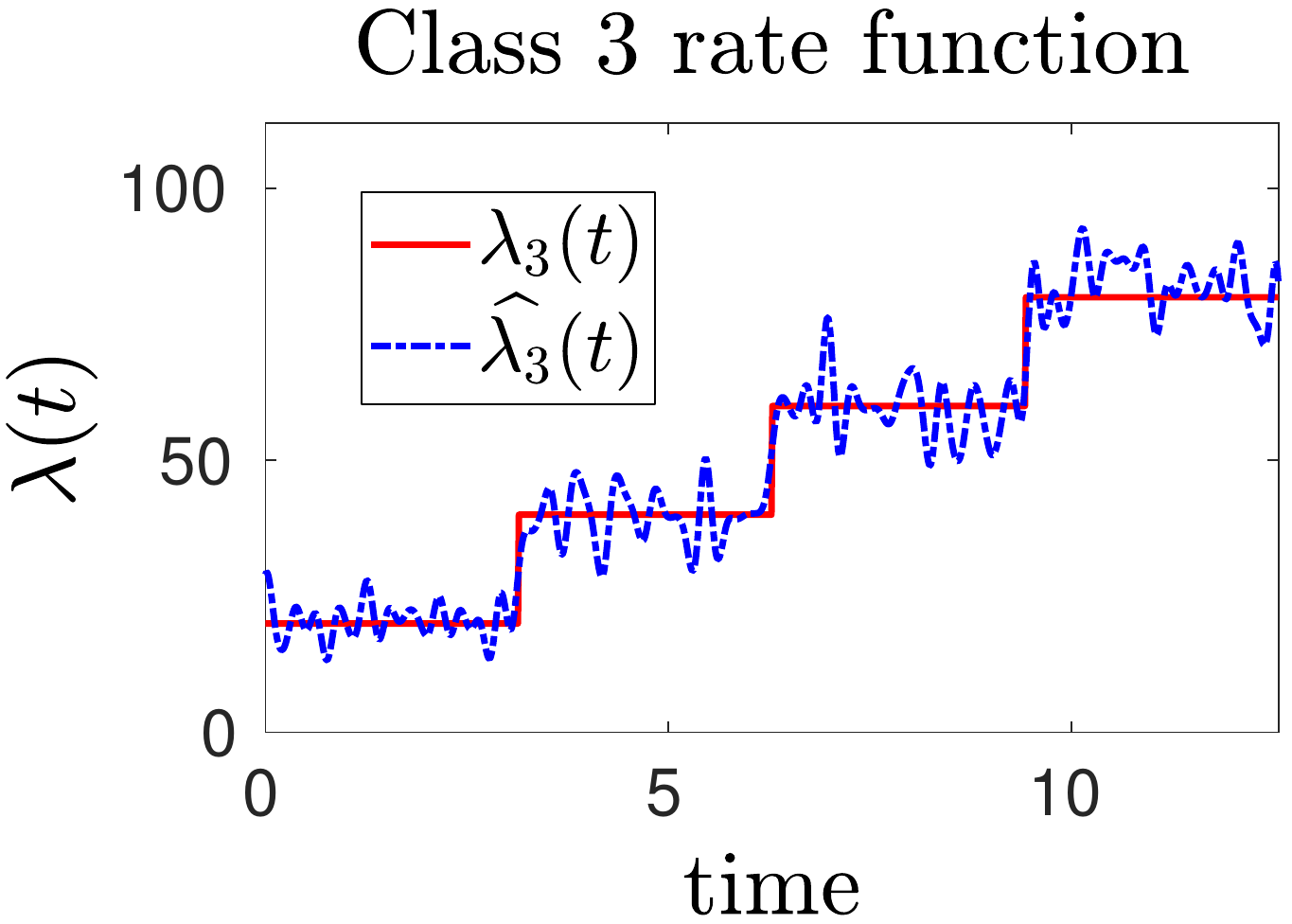}
\includegraphics[width=0.24\textwidth]{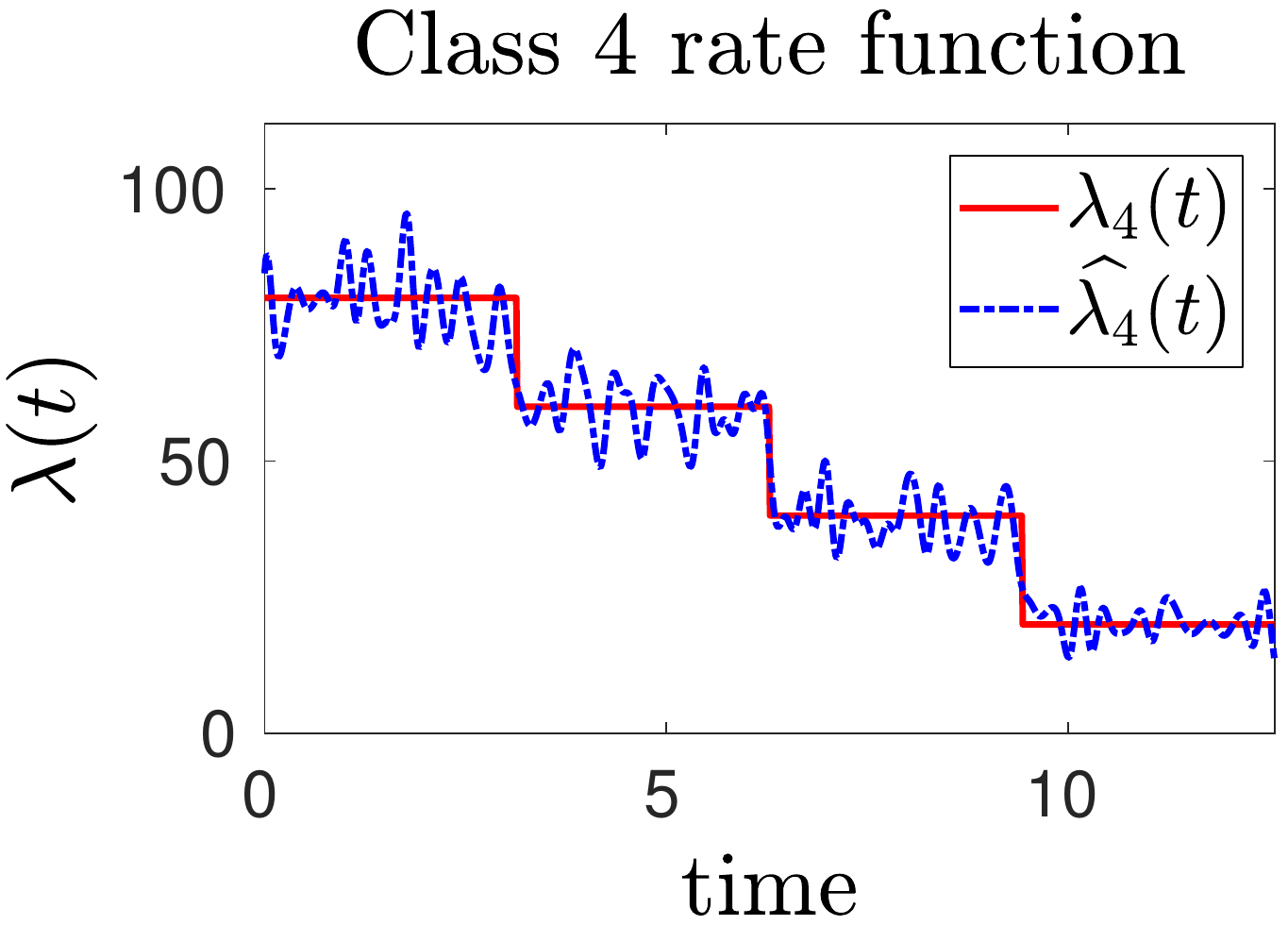}
\end{center}
\caption{Results of the synthetic data study detailed in Section \ref{sec:synthetic_results}. The top row of plots shows raster plots of a selection of event times generated for each of the four rate functions. Subsequent plots show that rate functions for each of the 3 classification and clustering tasks can be recovered using the procedures described in Sections \ref{sec:classification} and \ref{sec:clustering}. $\lambda_{1}(t)$ (\emph{resp.} $\lambda_{2}(t), \lambda_{3}(t), \lambda_{4}(t))$ is the prescribed rate functions for class 1 (2,3,4) and $\widehat{\lambda}_{1}(t)$ $\left(\widehat{\lambda}_{2}(t), \widehat{\lambda}_{3}(t), \widehat{\lambda}_{4}(t)\right) $is its estimate. 100 cubic basis functions (i.e. $n_{b}=100$) were used for the rate function basis expansions and all coefficients of these (i.e. $c_{1,1},\ldots, c_{1,100}, c_{2,1},\ldots, c_{2,100}, c_{3,1},\ldots, c_{3,100}, c_{4,1},\ldots, c_{4,100}$) had an initial value of 1 for the classification results.}
\label{fig:synthetic_plots}
\end{figure*}

Again, data for 20 observations was generated. Using the clustering method detailed in Section \ref{sec:clustering} we were able to recover the NHPP rate functions for all data sets (see Figure \ref{fig:synthetic_plots}). For each data set, we were able to correctly assign each observation to the NHPP model from which it's event time data was generated by assigning it to the model for which the membership probability (equation (\ref{eq:EM_mem_prob})) was maximal.

\subsection{Retail transaction data results}
Next we introduce our first real data set which is made up of the till transaction times from the 6$^{\mbox{th}}$ of September 2011 to the 28$^{\mbox{th}}$ of February 2015 for 74 UK stores (10,000 transactions per store) belonging to a large UK retailer. Half of these stores have been categorised by the retailer as `High street' (HS) which are large stores found in city centres and the remainder as `travel' stores which are situated in airports and railway stations. These categorisations serve as the ground truth for our analysis. 

\subsubsection{Classification}\label{sec:class_boots}
We used 5 fold cross validation (CV) \citep{efron1983leisurely} to obtain estimates for the classification accuracy, high street store and travel store true positive rates (TPR). We repeated the CV procedure 100 times with different random fold partitions to give a total of 500 numerical experiments. The average classification accuracy, high street store true positive and travel store true positive rates were 0.847, 0.994 and 0.701 respectively. The estimates for the rate functions of each class for one CV run are shown in Figure \ref{fig:boots}. Both rate functions are periodic with the high street rate function peaking around Christmas which corresponds to the busiest shopping period on the high street. In contrast, the rate function for the travel stores (which are located in airports and railway stations) peaks around summer which coincides with the peak holiday period when holiday makers are travelling via transportation hubs. For our results, the high street store true positive rate was higher than the travel store true positive rate suggesting the classifier has more of a tendency to misclassify travel stores rather than high street stores. Further investigation reveals that the misclassified travel stores are situated in central railways stations in large UK cities including London, Glasgow, Leeds and Liverpool. A possible explanation for the misclassification of these stores is that, although classed as travel stores, the store location means that these stores predominantly serve high street shoppers and thus have a demand profile which is similar to high street stores.

\subsubsection{Clustering}
\begin{figure}[htbp]
\begin{center}
\hspace{2cm} \textbf{Till transaction times} \hspace{2cm} \textbf{Classification results}\\
\includegraphics[width=0.32\textwidth]{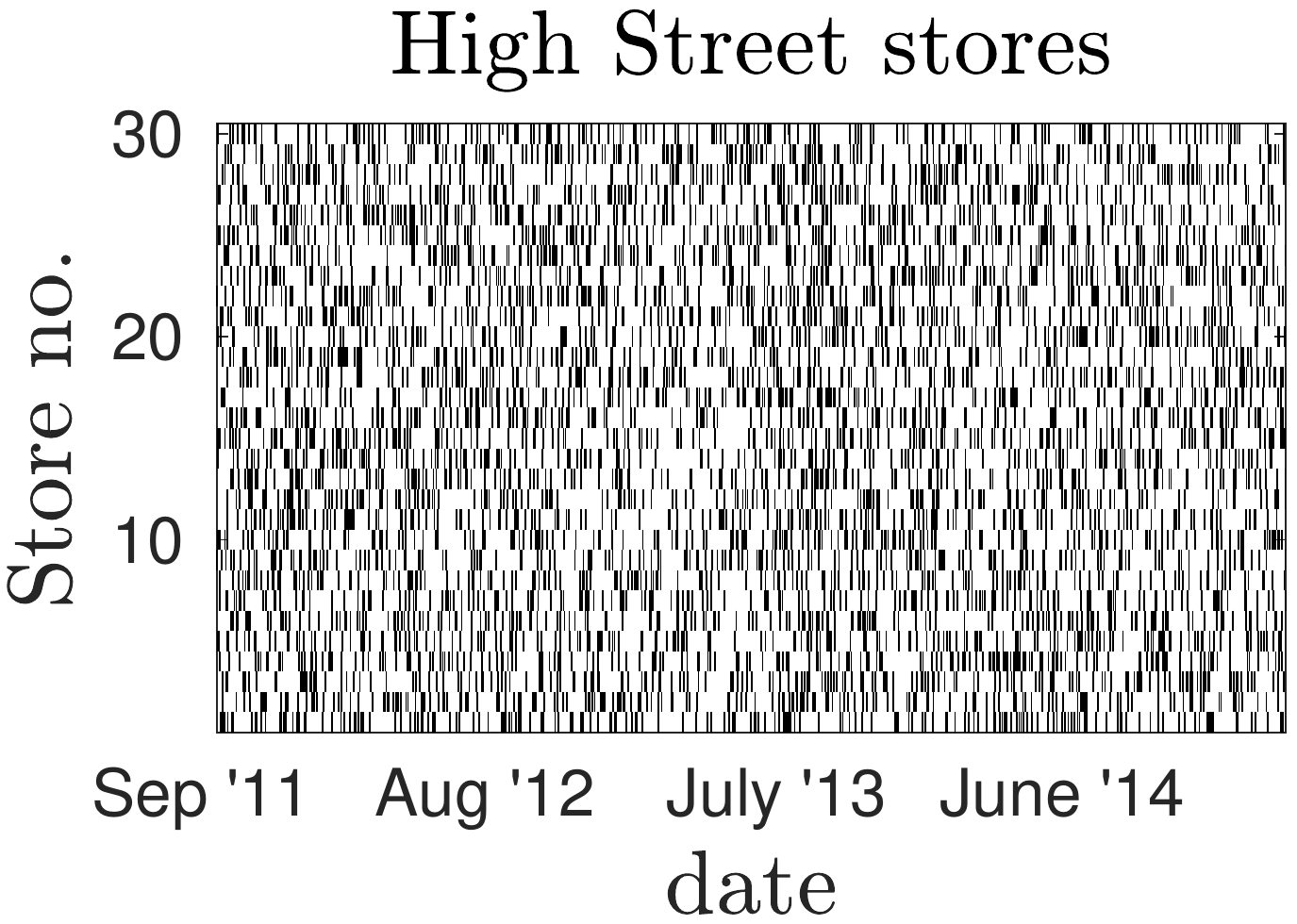}
\includegraphics[width=0.32\textwidth]{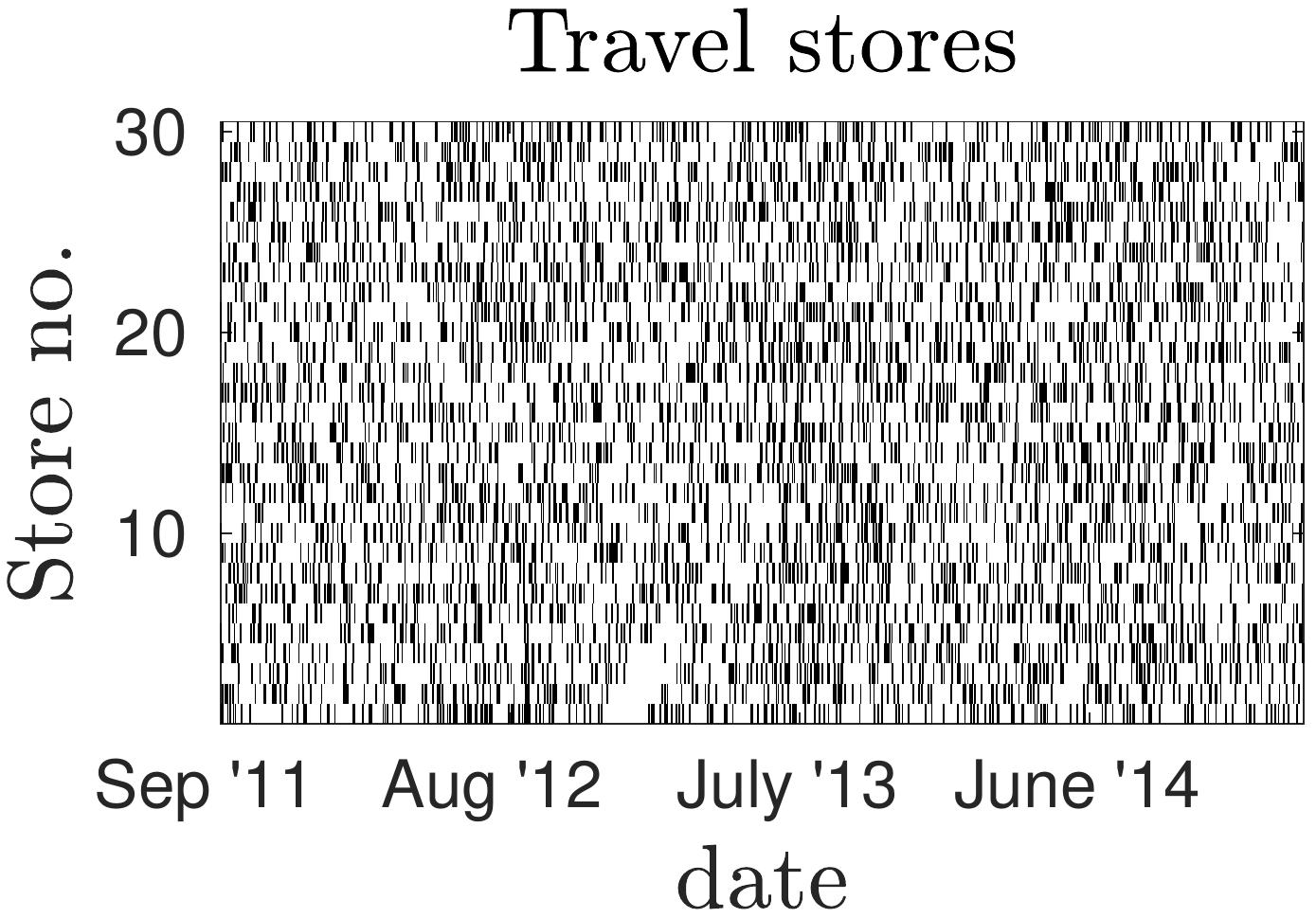} 
\includegraphics[width=0.32\textwidth]{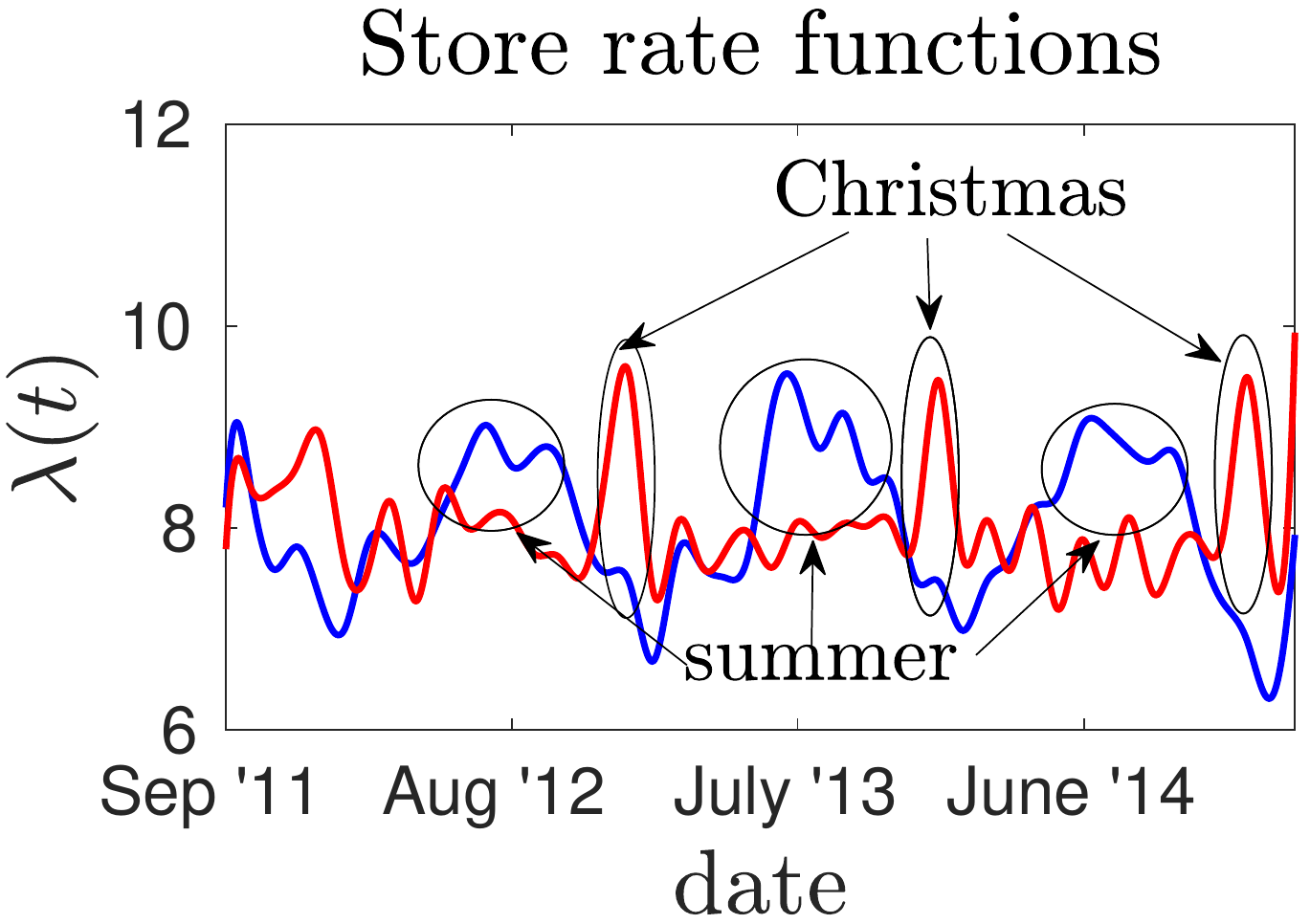} \\
\hspace{-3.5cm} \textbf{Clustering results}\\
\hspace{-4cm} \textbf{$\boldsymbol{k=2}$} \hspace{2.5cm} \textbf{$\boldsymbol{k=3}$}\\
\includegraphics[width=0.32\textwidth, valign=t]{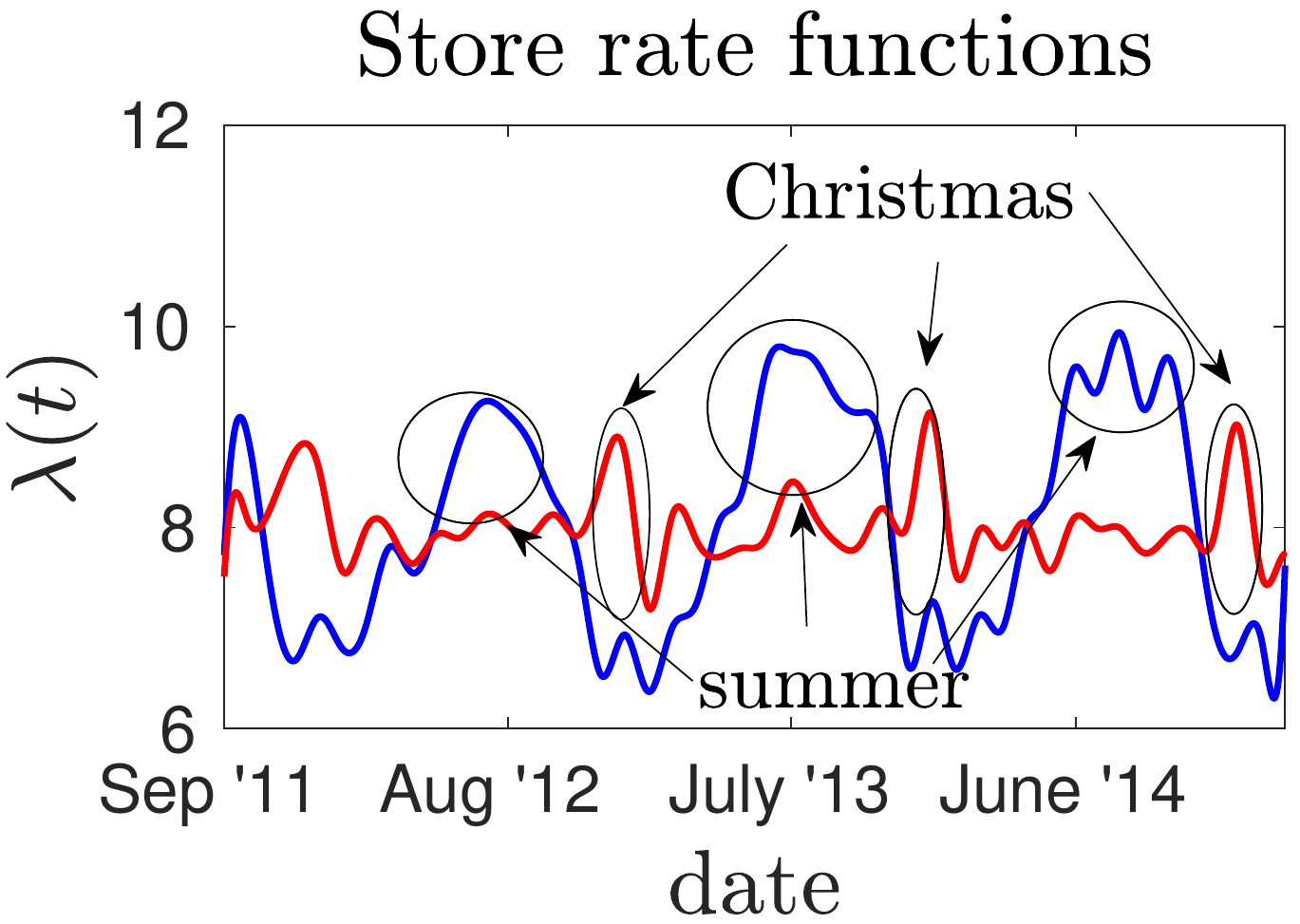}
\includegraphics[width=0.32\textwidth, valign=t]{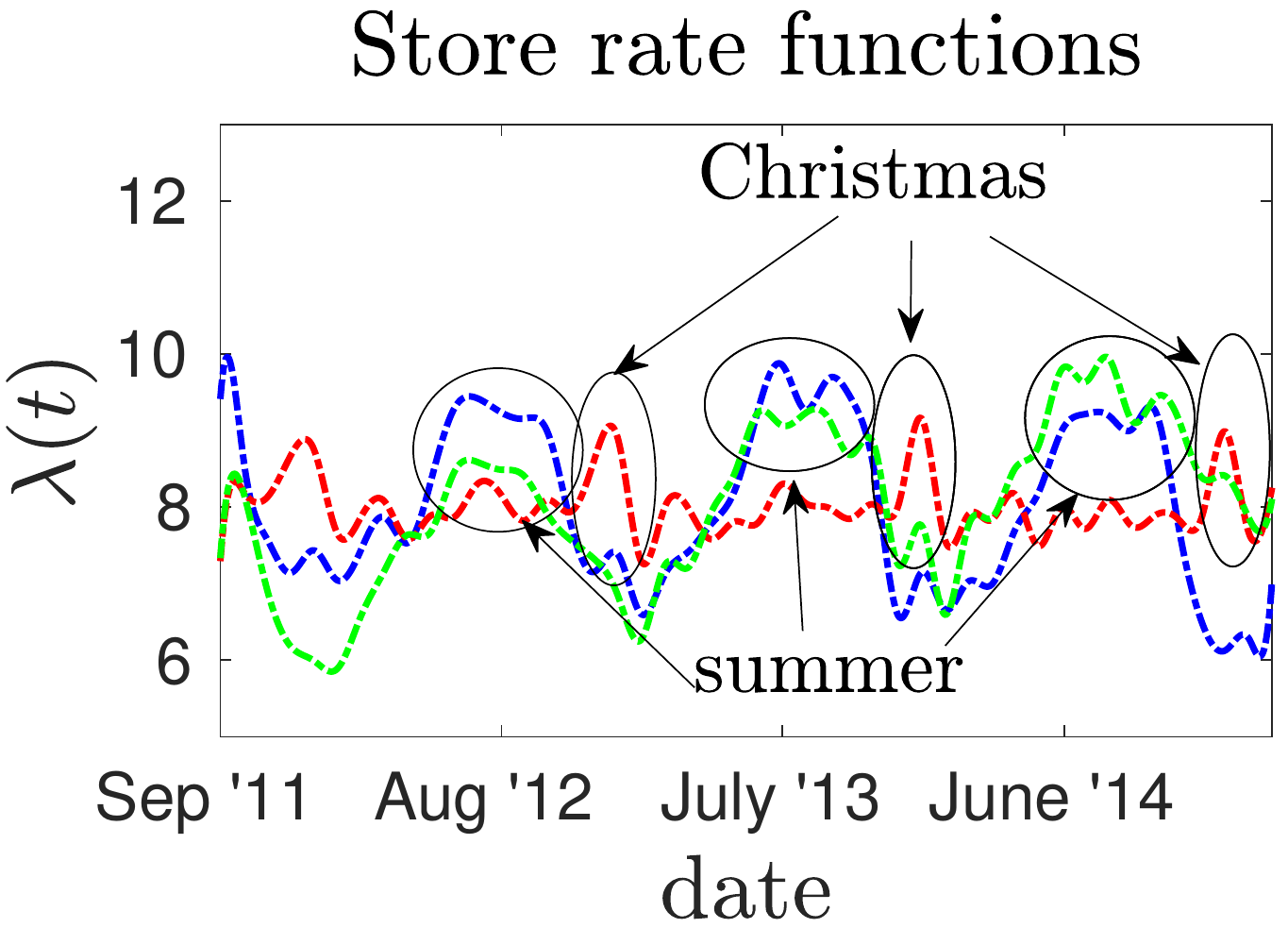} \hspace{0.2cm}
\includegraphics[width=0.3\textwidth, valign=t]{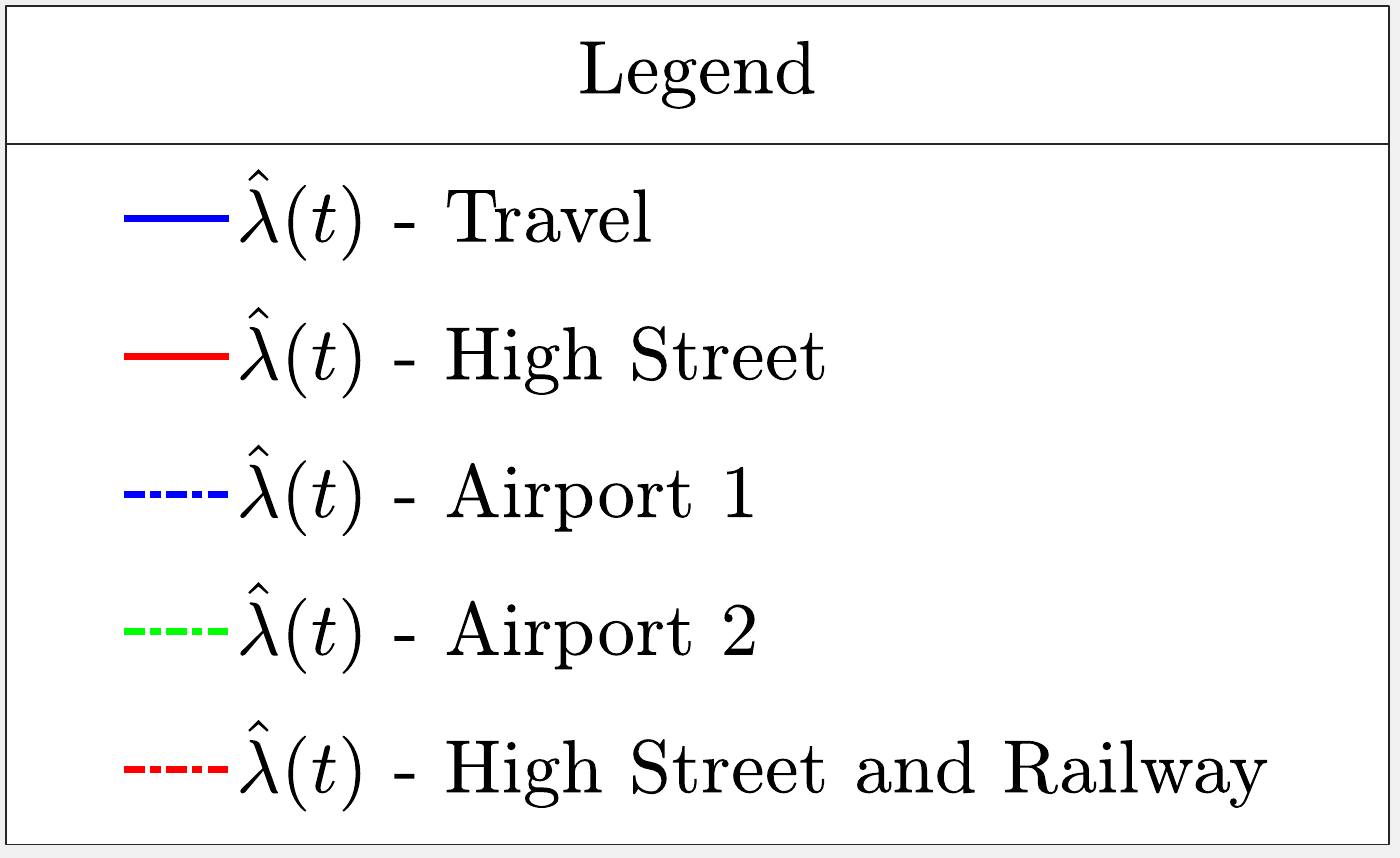}
\end{center}
\caption{Retail store results. The first row of plots include raster plots of till transaction event times for a selection of stores for each store class (`high street store' and `travel store' ). Also included within this row is a plot of the estimated rate functions for travel (blue) and high street (red) stores obtained from the classification procedure. These rate functions indicate peak demand for travel stores is around summer while demand peaks around Christmas for high streets stores. The final row of plots are the estimated rate functions obtained from the clustering method for a mixture of 2 and 3 NHPP models. For the clustering results for $k=2$ the blue rate function peaks around summer and the membership probabilities for most travel stores are greatest for this model. The red function peaks around Christmas and this rate function corresponds predominantly to high street stores.}
\label{fig:boots}
\end{figure}

Setting the number of models in the mixture to 2, and assigning each observation to the model for which its membership probability is maximal, the average clustering accuracy, high street store true positive and travel store true positive rates were 0.764, 1 and 0.528 respectively. The high street store true positive rate (\emph{resp}. travel store true positive rate) of the clustering results was higher (lower) than the corresponding classification results. Further investigation in to the cause of this discrepancy revealed that most of the stores in the `travel' cluster are located in airports. Stores categorised by the retailer as travel stores, but with temporal transaction characteristics more similar to high street stores (such as the stores situated in central railway stations in large UK cities mentioned in the previous section) cluster into the `high street' group. The rate functions of the two NHPP models are shown in Figure \ref{fig:boots} and clearly the blue rate function peaks around summer and corresponds to travel stores,while the red rate function peaks around Christmas and corresponds to high street stores.

Rate function estimates obtained by setting $k=3$ are also shown in Figure \ref{fig:boots}. Two clusters, with rate functions labelled `airport 1' and `airport 2' in the plot are comprised almost exclusively of airport stores and have very similar rate functions peaking around the summer. A third cluster, labelled `high street and railway' is made up mainly of high street stores and railway stores located in stations in city centres and its rate function peaks around Christmas. The fact that setting $k=3$ does not lead to more meaningful store partitions, compared to setting $k=2$, supports the notion that the stores fall naturally in to one of two clusters, the first made up predominantly of airport stores and the second of stores located centrally in cities either on the high street or in city centre railway stations.

\subsection{Hubway bike share data set}\label{sec:hubway}
This data set contains bike trip information for the Hubway bike share program located in Boston, Massachusetts and its environs. The data, which covers the period from the 28$^{\text{th}}$ of July 2011 to the 30$^{\text{th}}$ of November 2013, contains trip duration time, date and time of the trip as well the the id of the start and finishing bike share stations of cycle journeys. Additionally the user type (\emph{casual} - `24-hour or 72-hour pass user' or \emph{member} - `annual or monthly member') was also recorded. In our analysis we ranked each station according to the proportion of station users classed as either \emph{casual} or \emph{members} and classified the $\kappa$ stations with the highest casual-member proportions as `casual' stations and the $\kappa$ stations with the lowest casual-member proportions as `member' stations. We only considered stations with more than 5,000 recorded trips to avoid any spurious results caused by stations with only a handful of recorded trips. This left a total of 89 station observations. The times of day at which the bike trips began at each station served as the sample of event times for each station observation. 

\subsubsection{Classification}
We used the same CV procedure as before to obtain average  classification accuracy, casual station and members station true positive rates. Raster plots showing the time of day at which bike trips started for both station types and plots showing the average performance of our classification procedure over all test folds as a function of $\kappa$ are shown in Figure \ref{fig:hubway}. The results show that our classification procedure did a very good job of classifying the station observations (the average classification accuracy for $\kappa=20$ is 0.775 for example). We also include a plot showing the estimated rate functions for the casual and member station classes for $\kappa=20$ in Figure \ref{fig:hubway}. The member stations rate function (red curve) has two distinct peaks around 8\textsc{am} and 5\textsc{pm}. These peaks are far less pronounced for the casual stations (blue curve) which has a larger intensity in the middle of the day. The most plausible explanation for these features is that member stations are predominantly used by commuters who cycle to and from work and, as they use the bike scheme every day, are more likely to sign up for membership. Whilst casual stations are more likely to be used by tourists who have the free time to cycle during the middle of the day but, as they use the bike share scheme infrequently, are less likely to become members.

Although the classification results are very good, for the 2 class problem, our classifier had a tendency to misclassify member stations more than casual stations. In Figure \ref{fig:hubway} is shown the estimated rate functions for the two `member stations' which were misclassified the most number of times during the CV experiments. Rather than two distinct peaks around 8\textsc{am} and 5\textsc{pm}, station 9 has a single peak around 8\textsc{am} and station 15 a single peak around 5\textsc{pm}. This suggests that there may be two distinct sub-types of member stations; one type used mostly in the mornings (most likely located in the suburbs close to the homes of commuters) and another close to the commuters places of work (most likely in the city centre) which are mostly used after 5\textsc{pm} when the commuters return home from work. This motivated us to consider a three class problem consisting of `casual stations', `morning member stations' (member stations where the majority of trips took place before 5\textsc{pm}) and `afternoon member stations' (member stations where most trips took place after 5\textsc{pm}). The results of this experiment are also shown in Figure \ref{fig:hubway} where it can be seen that the average classification accuracy for the 3 class problem is higher than for the 2 class problem (for $\kappa=20$, the 3 class problem average classification accuracy is 0.829, whilst it is 0.775 for the 2 class problem for example). Furthermore, the rate function estimates for the 3 class problem, which are also shown in Figure \ref{fig:hubway}, capture the different usage patterns of the 3 different station types for the 3 class problem. There is a clear peak in activity for the morning (\emph{resp.} afternoon) member station class around 8\textsc{am} (5\textsc{pm}), whilst the estimated rate function for the casual station type has more of a uniform profile.

\subsubsection{Clustering}
\begin{figure}
\hspace{4cm} \textbf{Bike trip start times}\\
\floatbox[{\capbeside\thisfloatsetup{capbesideposition={left,top},capbesidewidth=4cm}}]{figure}[\FBwidth]
{\caption{Classification and clustering results for $k=2$ (`member' and `casual' station classes) and 3 (`morning (AM) member', `afternoon (PM) member' and `casual' station classes) for the Hubway bike share data. Plots show the performance of the classification and clustering methods as a function of $\kappa$ (see text for details of this parameter) and estimated rate functions for the station classes for $\kappa=20$. Estimated rate functions for stations with ids 9 and 15, which were the most misclassified `member' stations for $k=2$, are also shown.}\label{fig:hubway}}
{\includegraphics[width=0.32\textwidth]{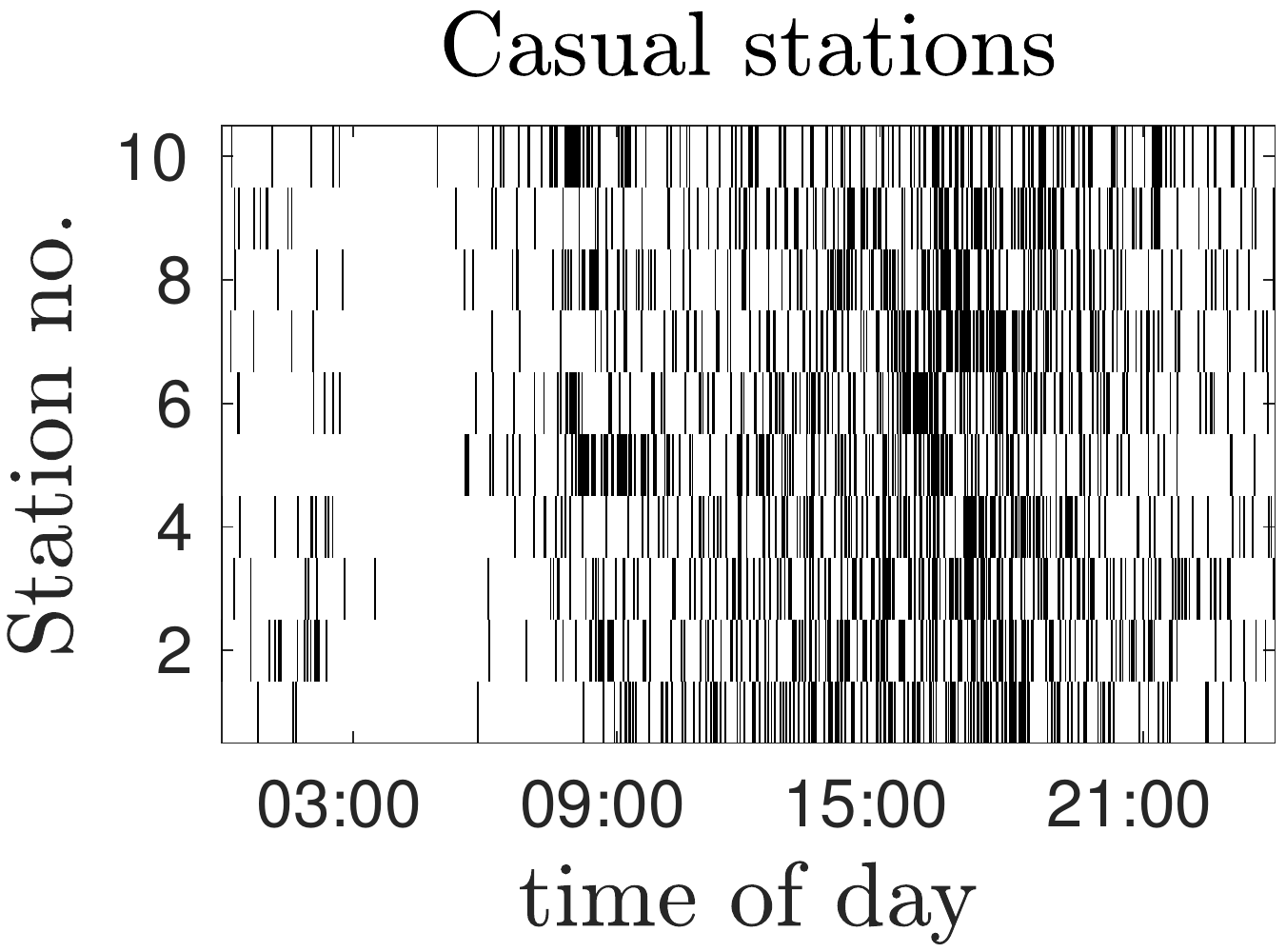}
\includegraphics[width=0.32\textwidth]{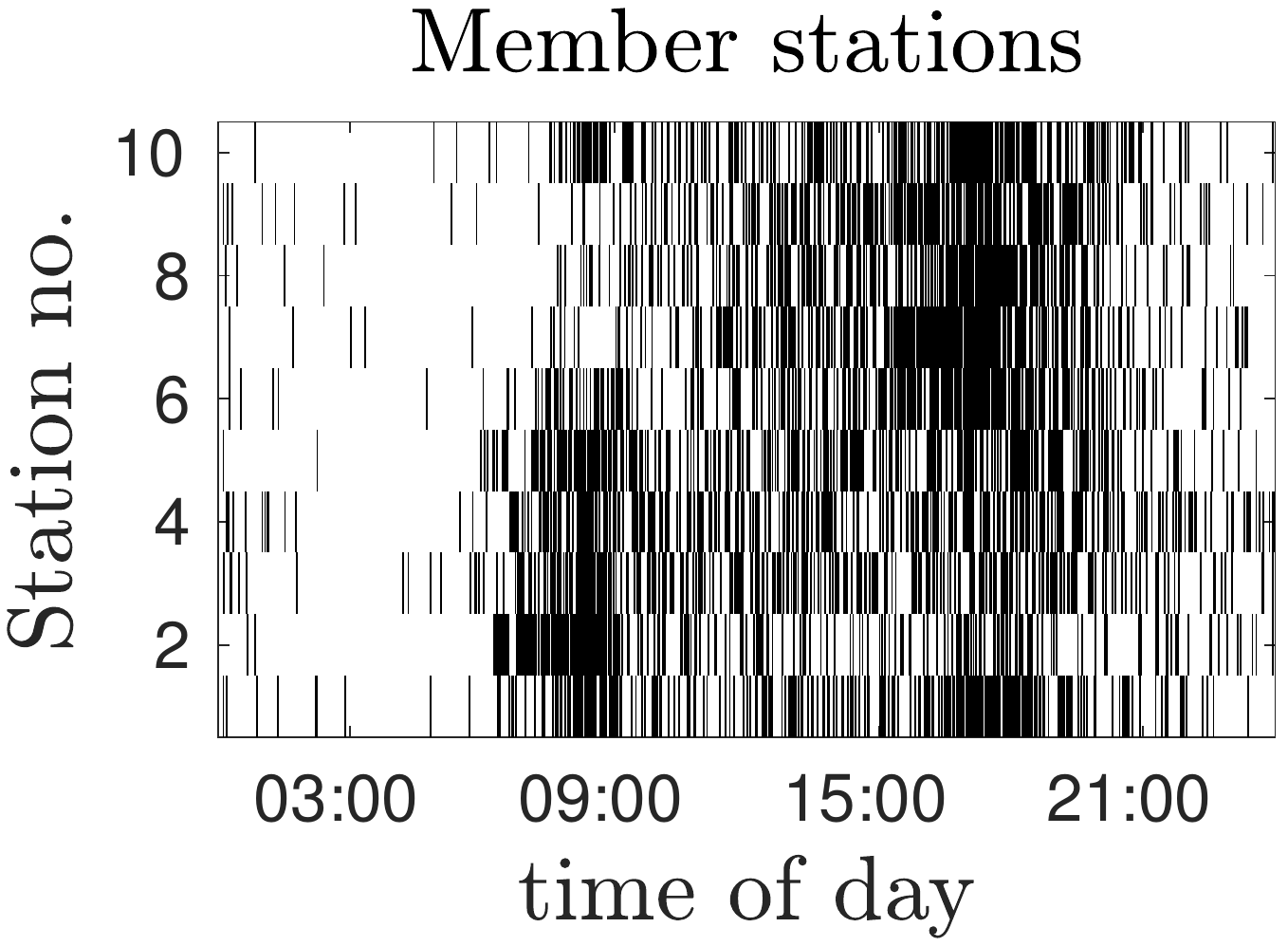}}
\vspace{-3.3cm}
\hspace{4cm} \textbf{Classification results for $\boldsymbol{k=2}$}\\
\hspace{4.15cm} \includegraphics[width=0.32\textwidth]{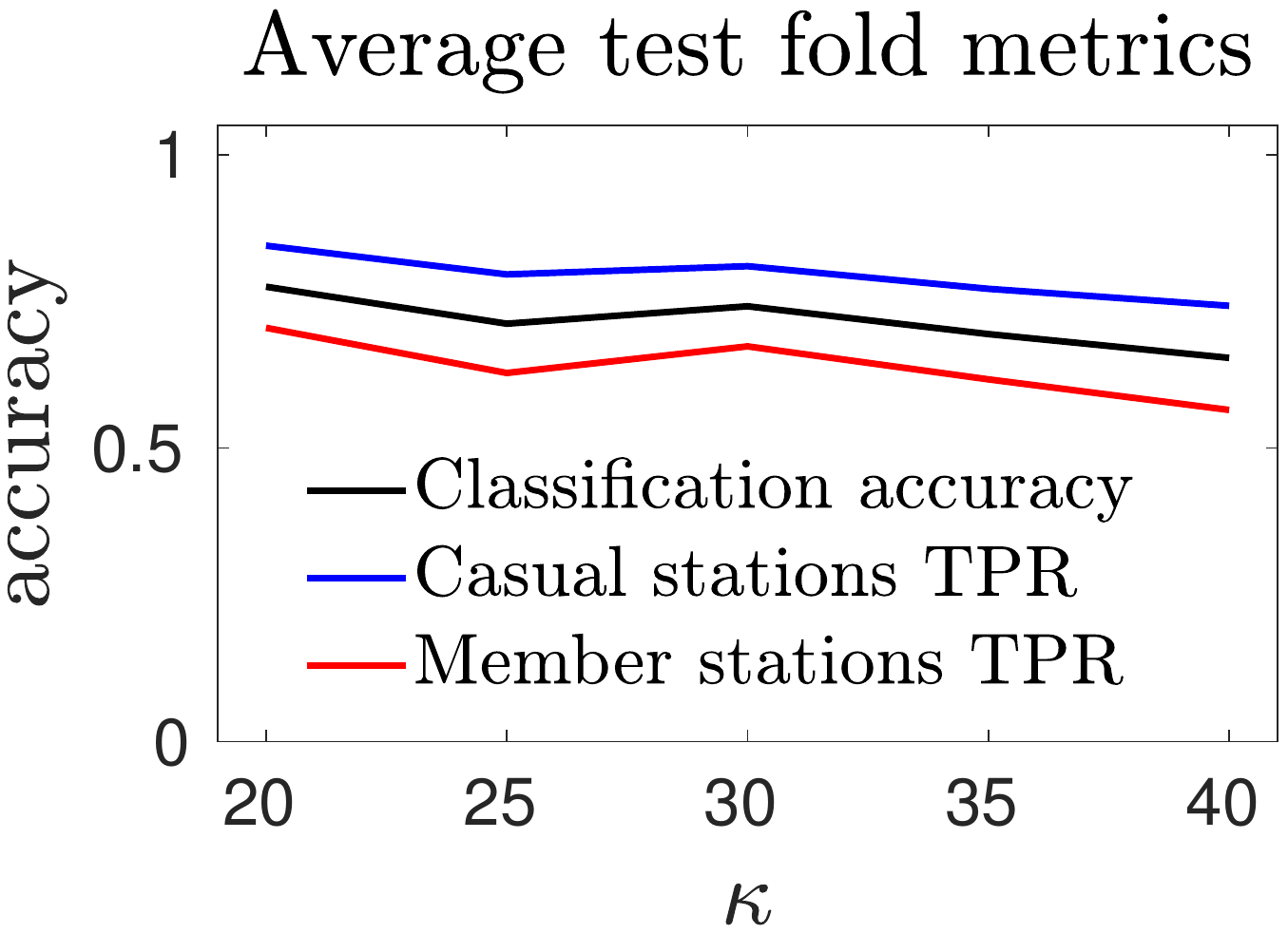}
\includegraphics[width=0.32\textwidth]{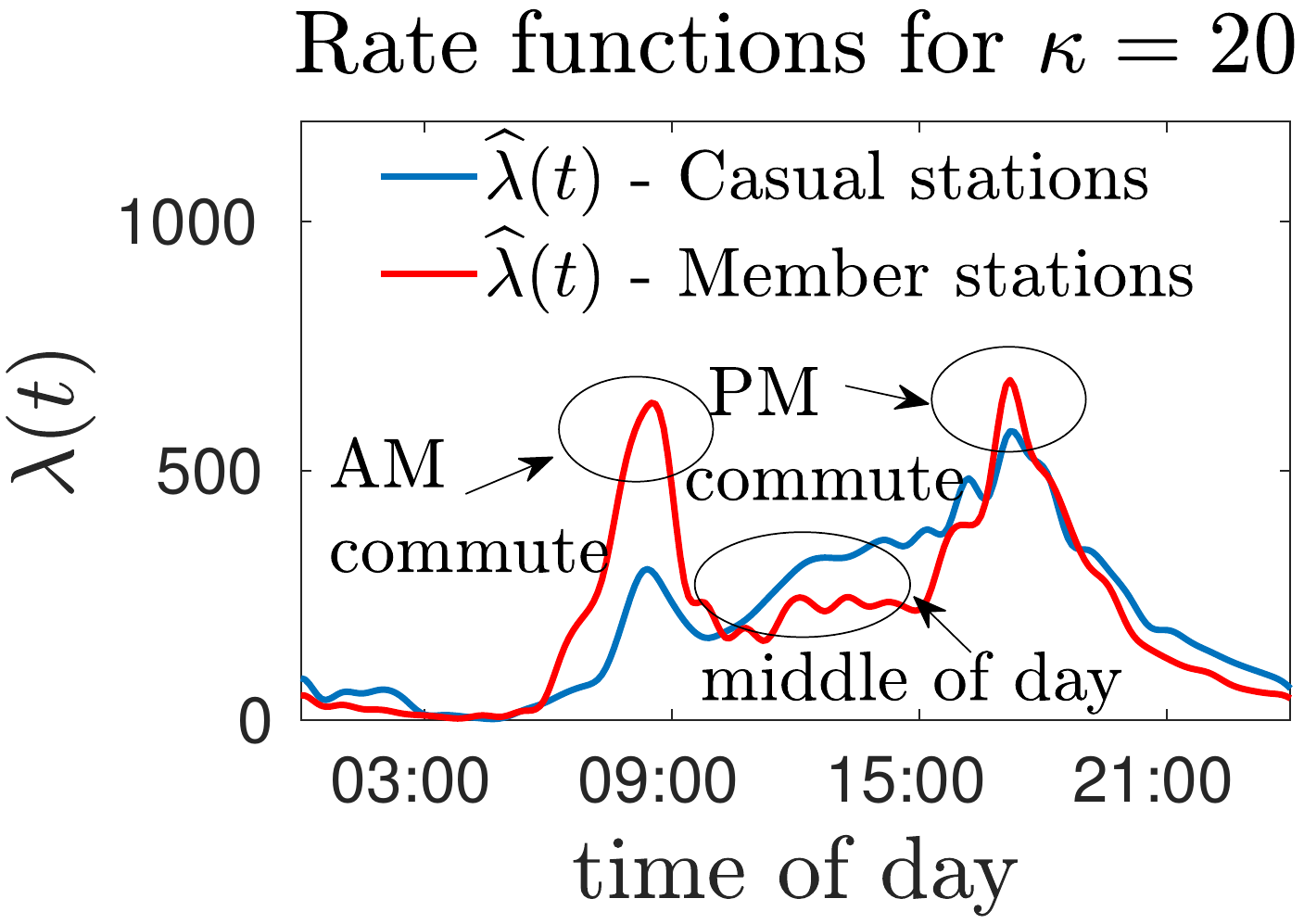}\\
\hspace{4.3cm} \textbf{Misclassified `member' stations for $\boldsymbol{k=2}$}\\
\hspace{4.15cm} \includegraphics[width=0.32\textwidth]{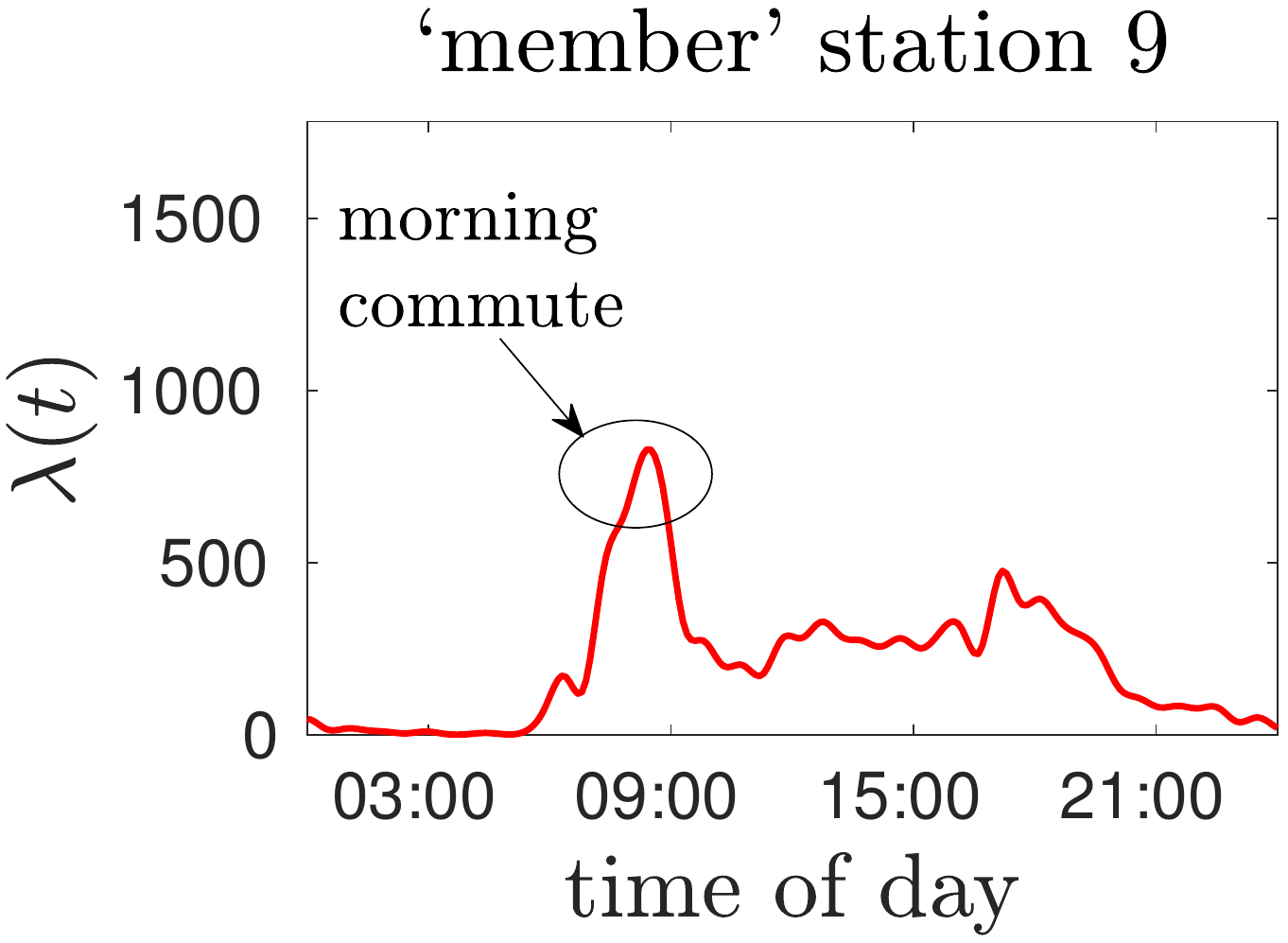}
\includegraphics[width=0.32\textwidth]{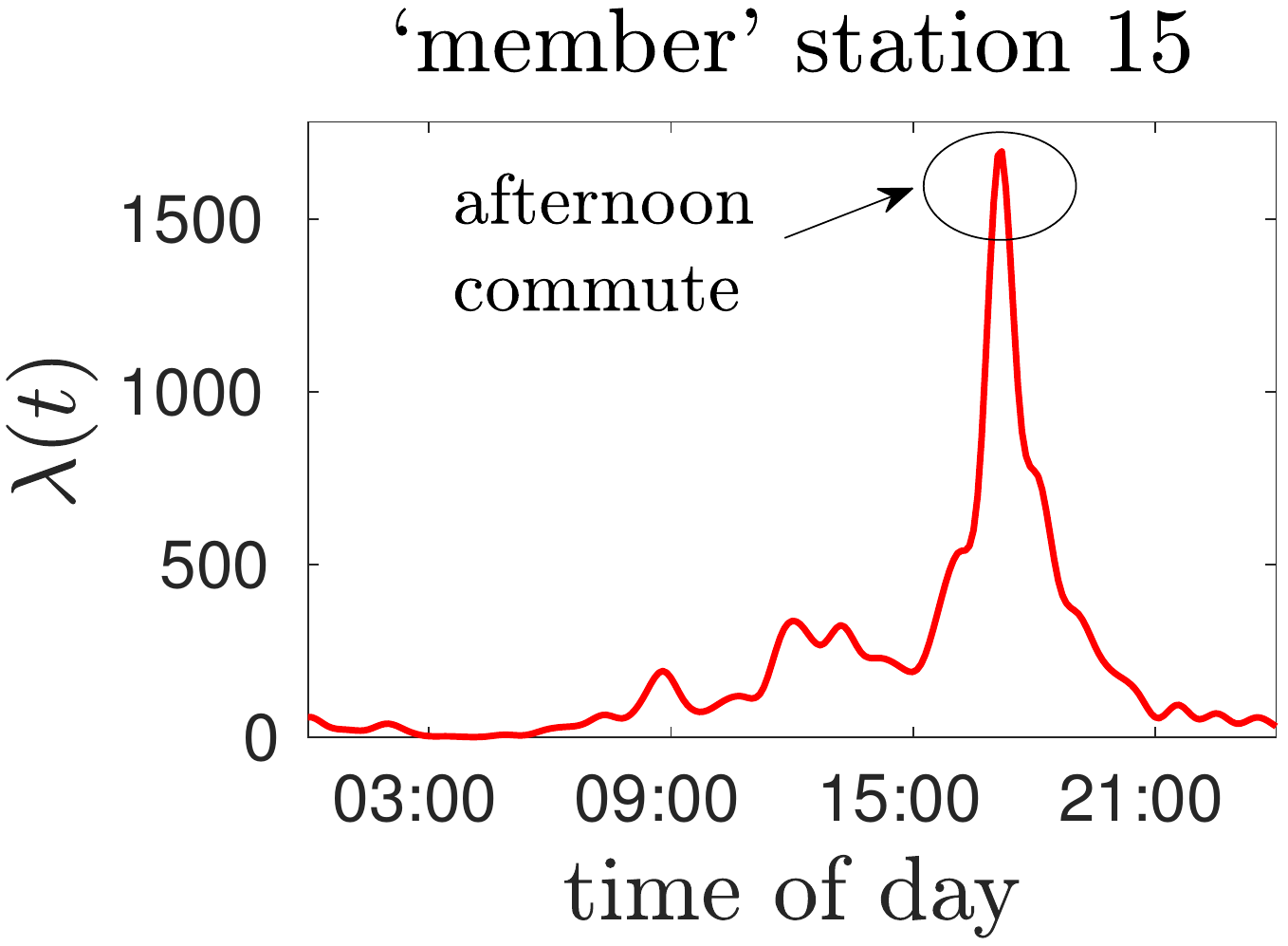}\\
\hspace{4cm} \textbf{Classification results for $\boldsymbol{k=3}$}\\
\hspace{4.15cm} \includegraphics[width=0.32\textwidth]{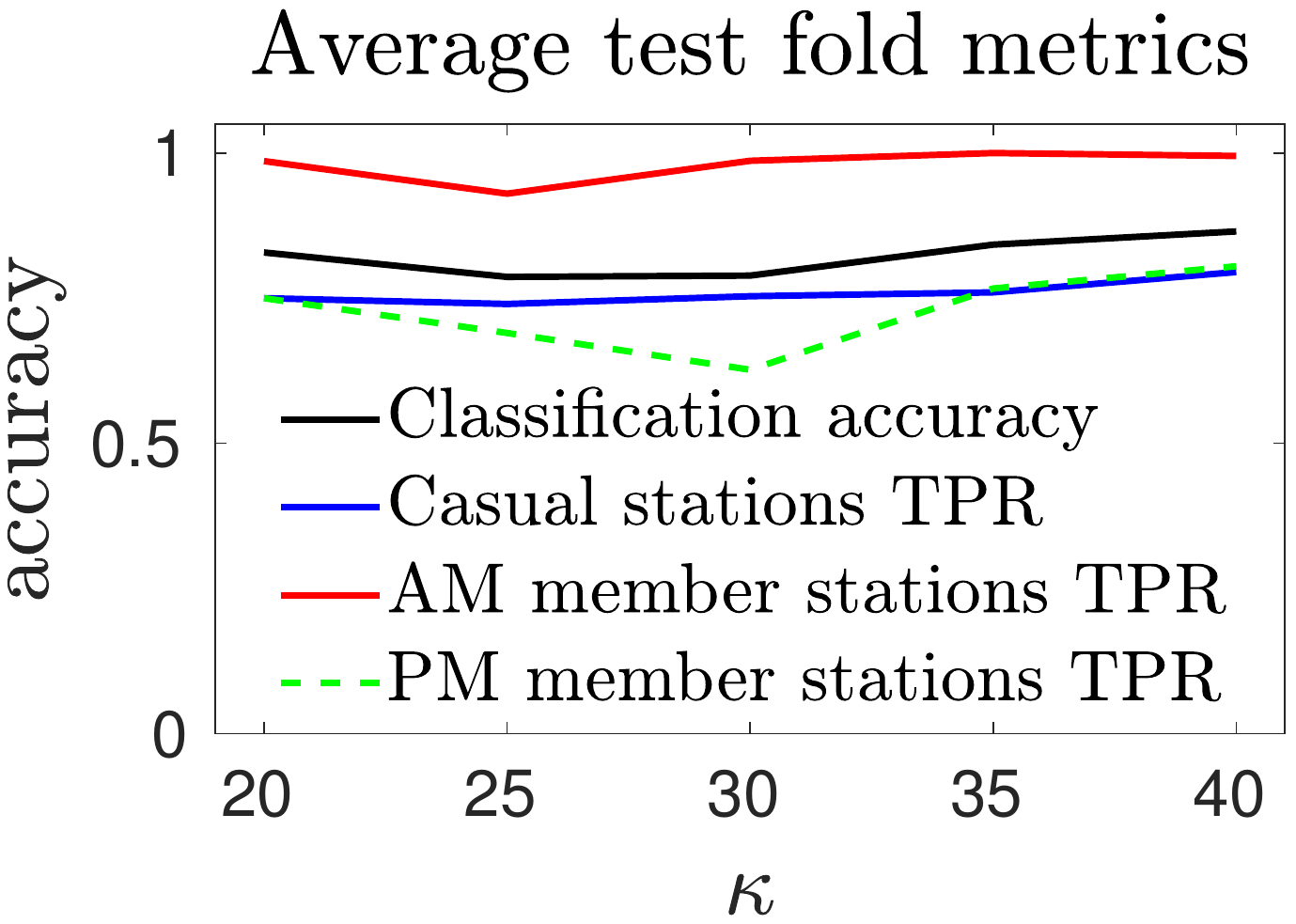}
\includegraphics[width=0.32\textwidth]{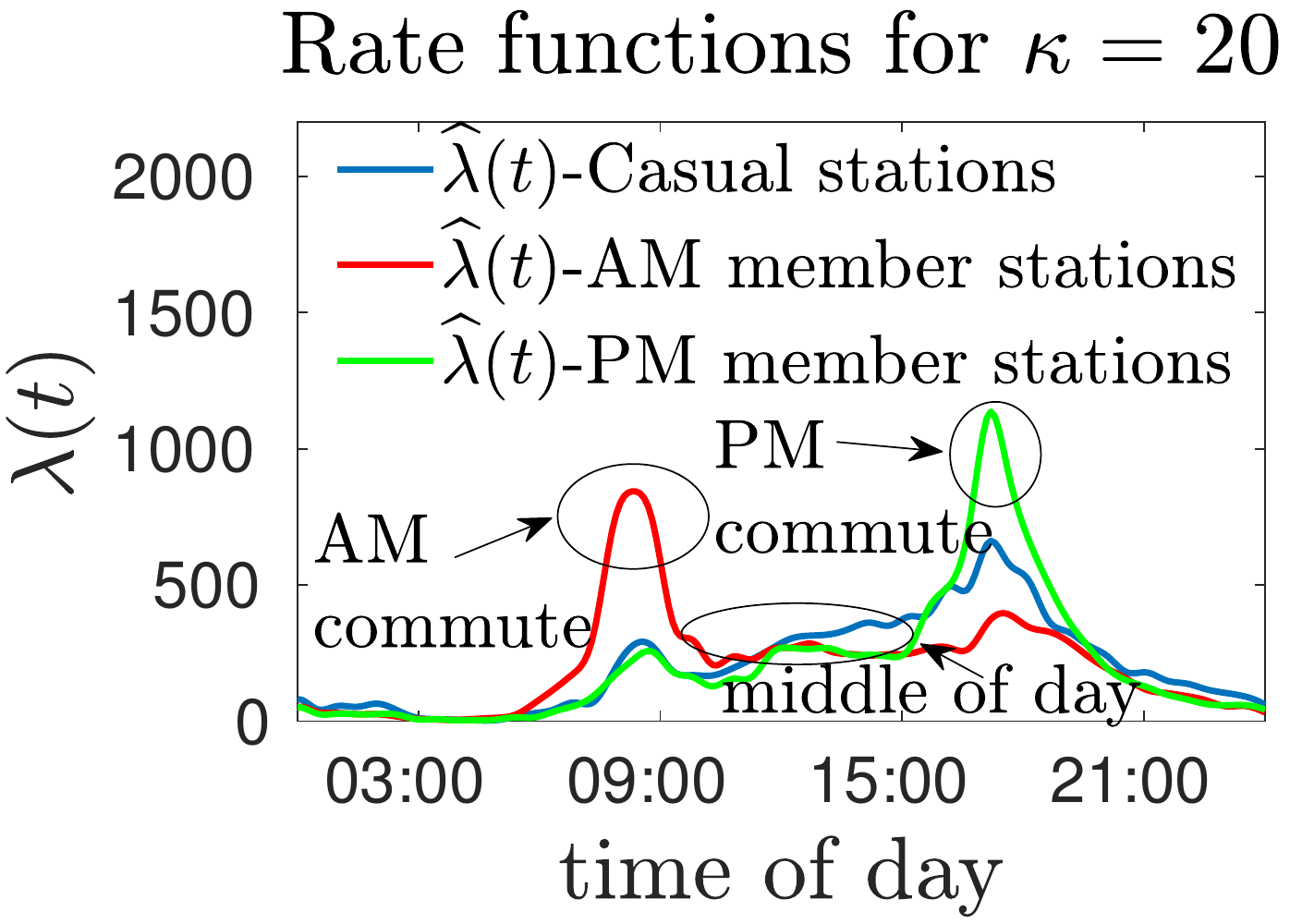}\\
\hspace{4cm}\textbf{Clustering results for $\boldsymbol{k=2}$}\\
\hspace{4.15cm} \includegraphics[width=0.32\textwidth]{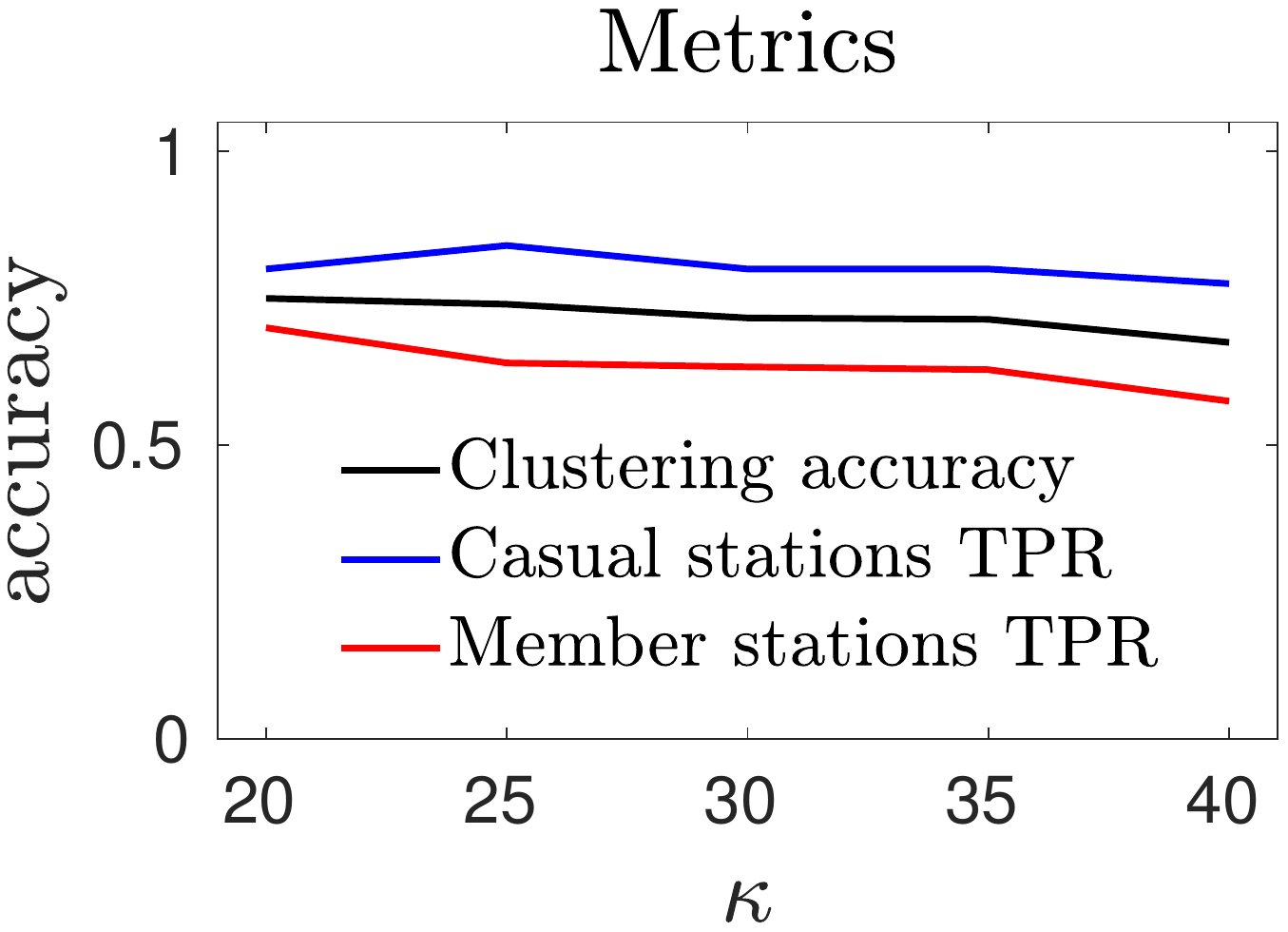}
\includegraphics[width=0.32\textwidth]{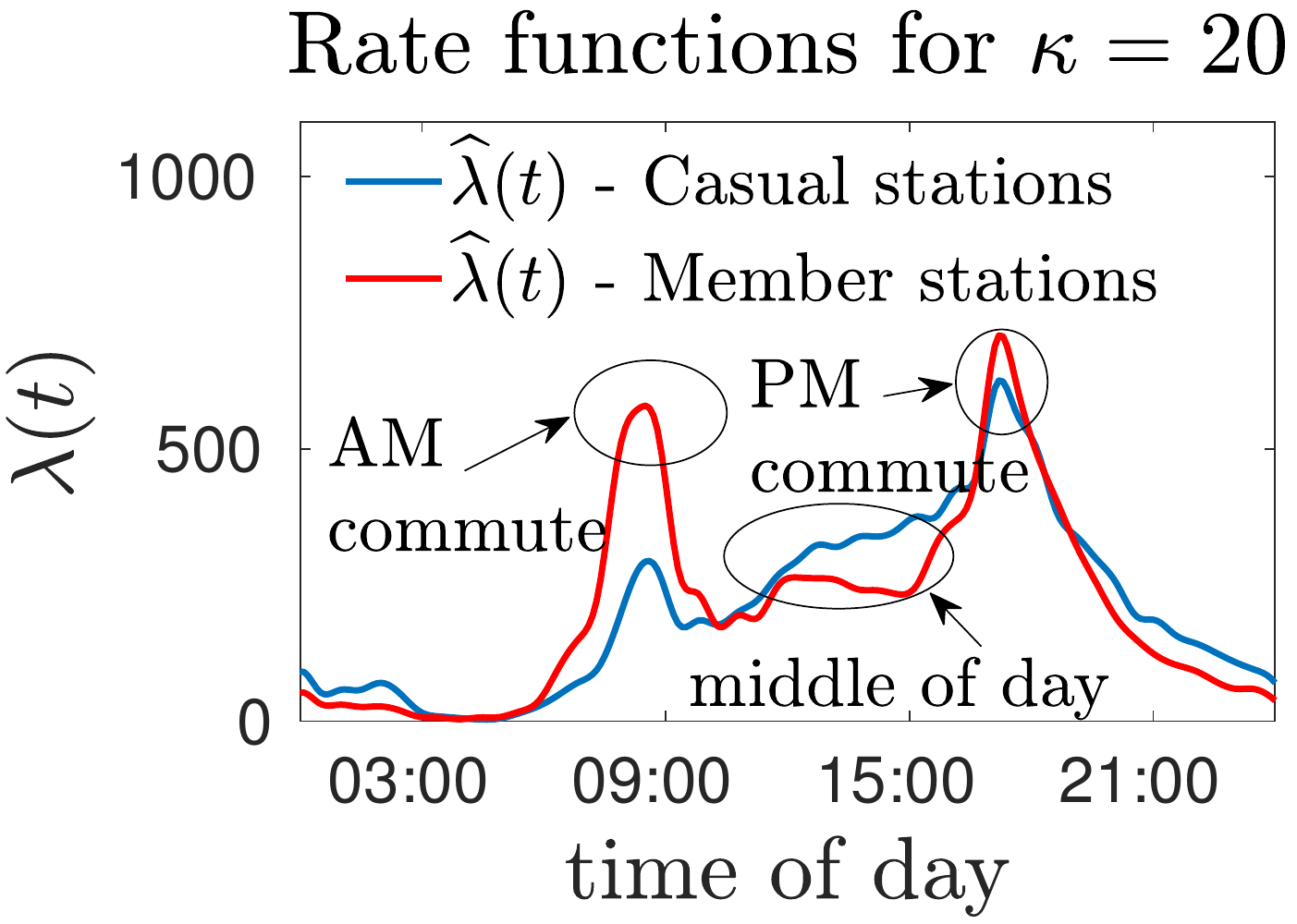}\\
\hspace{4cm}\textbf{Clustering results for $\boldsymbol{k=3}$}\\
\hspace{4.15cm}\includegraphics[width=0.32\textwidth]{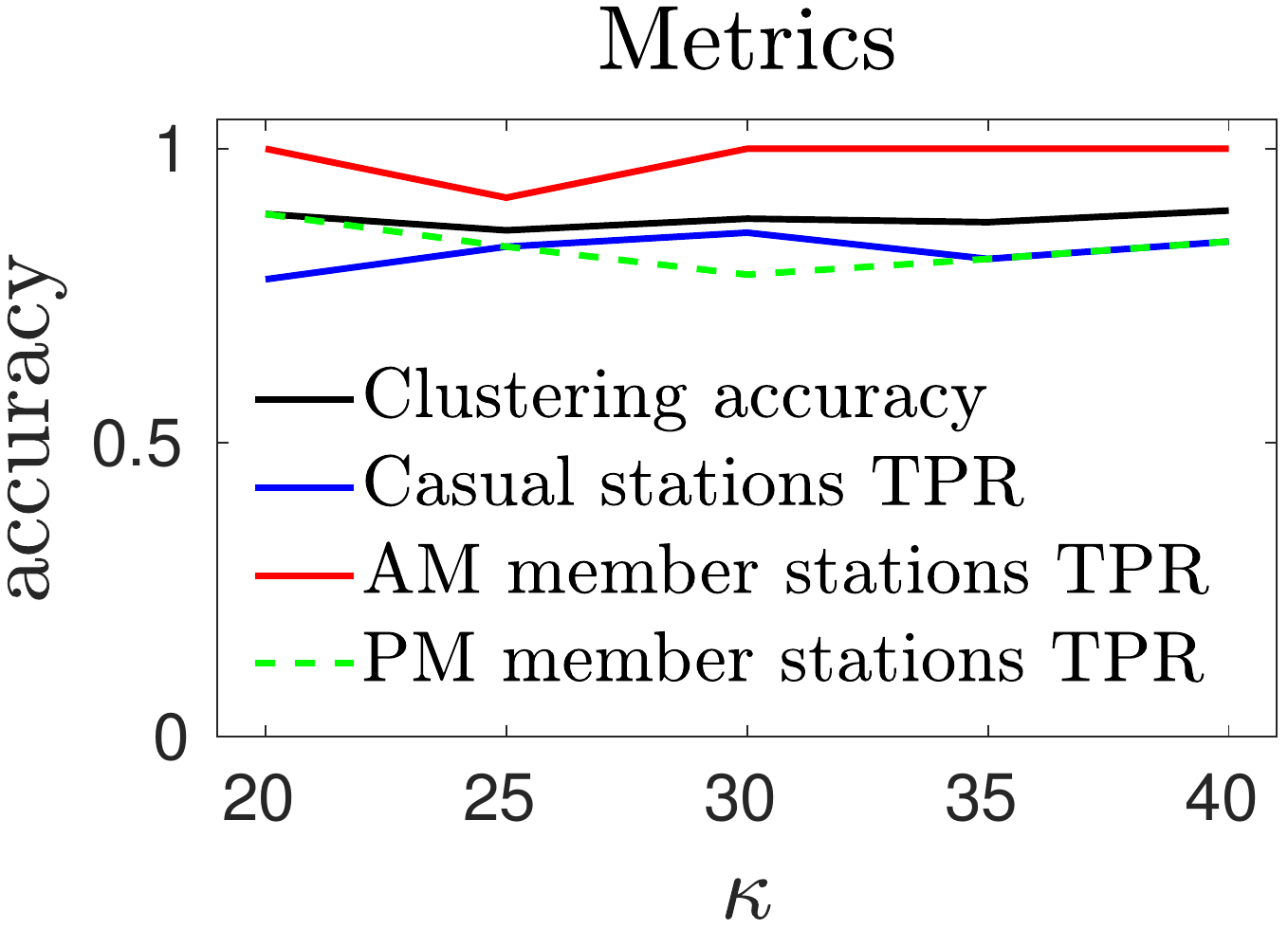}
\includegraphics[width=0.32\textwidth]{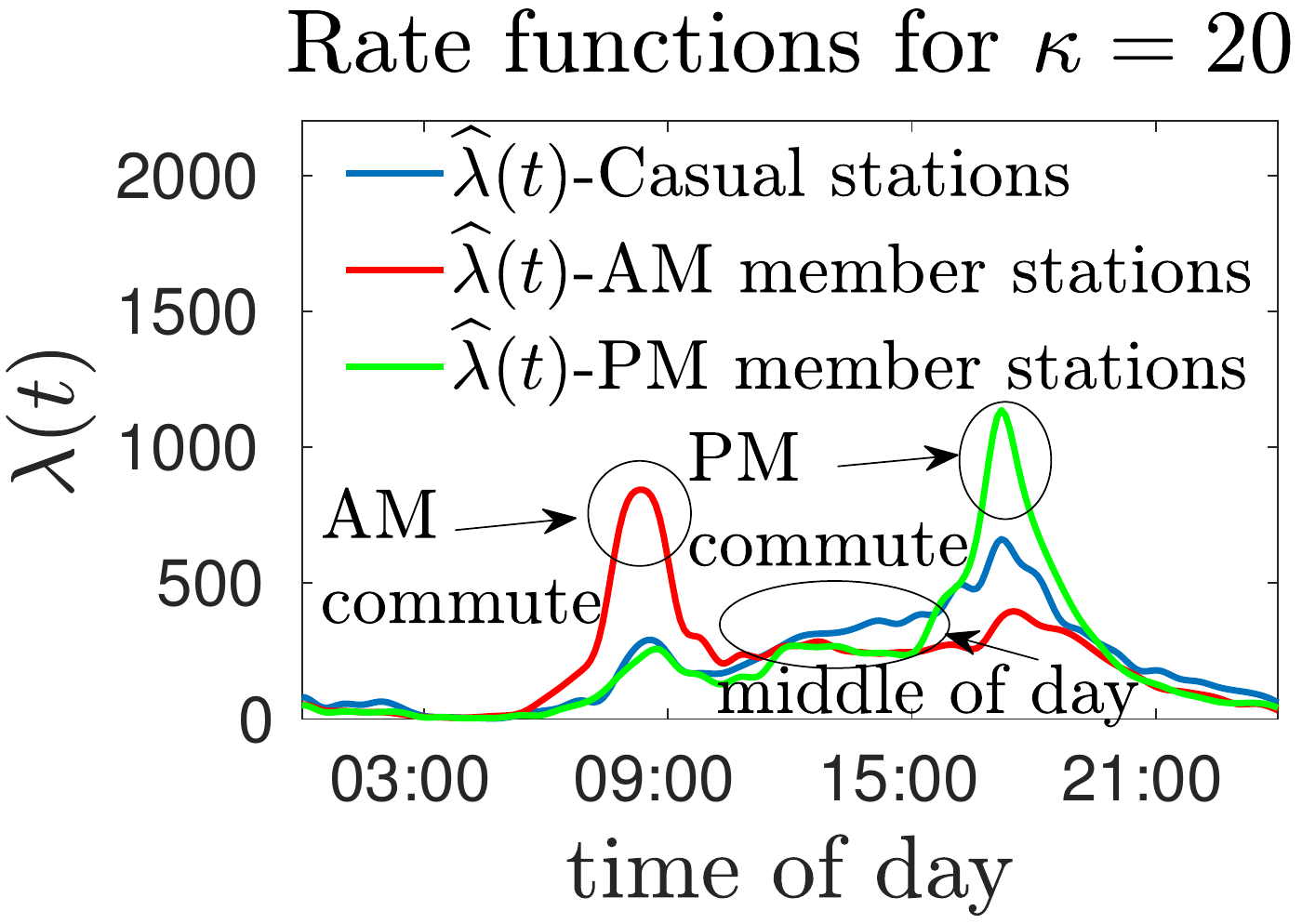}
\end{figure}

The clustering results for the 2 and 3 class problems are shown in Figure \ref{fig:hubway} and show that the stations classes defined in the previous subsection emerge naturally from the data. For $k=2$, the estimated rate function for one station type has two distinct peaks at 8\textsc{am} and 5\textsc{pm}. This is the `member station' class. The other estimated rate function has a greater intensity during the middle of the day. This is the `casual station' type, although for the clustering results, it has a noticeably higher peak around 5\textsc{pm} when compared to the rate function of the casual station class obtained via the classification procedure for $k=2$. For $k=3$, the casual station class persists and the two morning and afternoon member station sub classes, which were defined in the previous subsection, emerge naturally from the data. Assigning each observation to the class for which its membership probability is maximal, the clustering accuracy and true positive rates were very good (for $\kappa=20$, the clustering accuracy is 0.750 for $k=2$ and 0.889 for $k=3$) and broadly reflect the results seen for the classification task. For $k=2$, the casual station true positive rate exceeds that of the member station true positive rate. However, setting $k=3$ goes some way to remedying this discrepancy and also results in a higher overall clustering accuracy, as it did or the classification results. This indicates a choice of 3, rather than 2, clusters is a more appropriate choice for the clustering problem. 

\subsection{A\&E data set}\label{sec:AE}
This data set contains the time of day (00:00:00 to 23:59:59) in every month from April 2011 to the end of December 2014 at which patients arrived at Accident and Emergency (A\&E) departments in hospitals throughout England. Each patient was diagnosed as either suffering from poisoning (including alcohol poisoning) or a cardiac condition and these are used as the ground truths for the class labels. For each label there are 45 observations corresponding to each of the 45 months spanning April 2011 to the end of December 2014. Each monthly observation contains 10,000 A\&E admission events.

\subsubsection{Classification}
Using the same CV procedure as outlined in Section \ref{sec:class_boots}, we obtained an average  classification and true positive rates for both classes of 1. The estimated rate functions are shown in Figure \ref{fig:AE} from which it can be seen that admissions for poisoning peak around midnight. This is most likely due to instances of alcohol poisoning amongst revellers in the late evenings. Cardiac conditions peak in the morning, which is consistent with evidence in the medical literature \citep{peters1989propranolol, muller1989circadian}.

\subsubsection{Clustering}
\begin{figure}[htbp]
\centering
\hspace{0.5cm} \textbf{Hospital admissions times}\\ 
\includegraphics[width=0.32\textwidth]{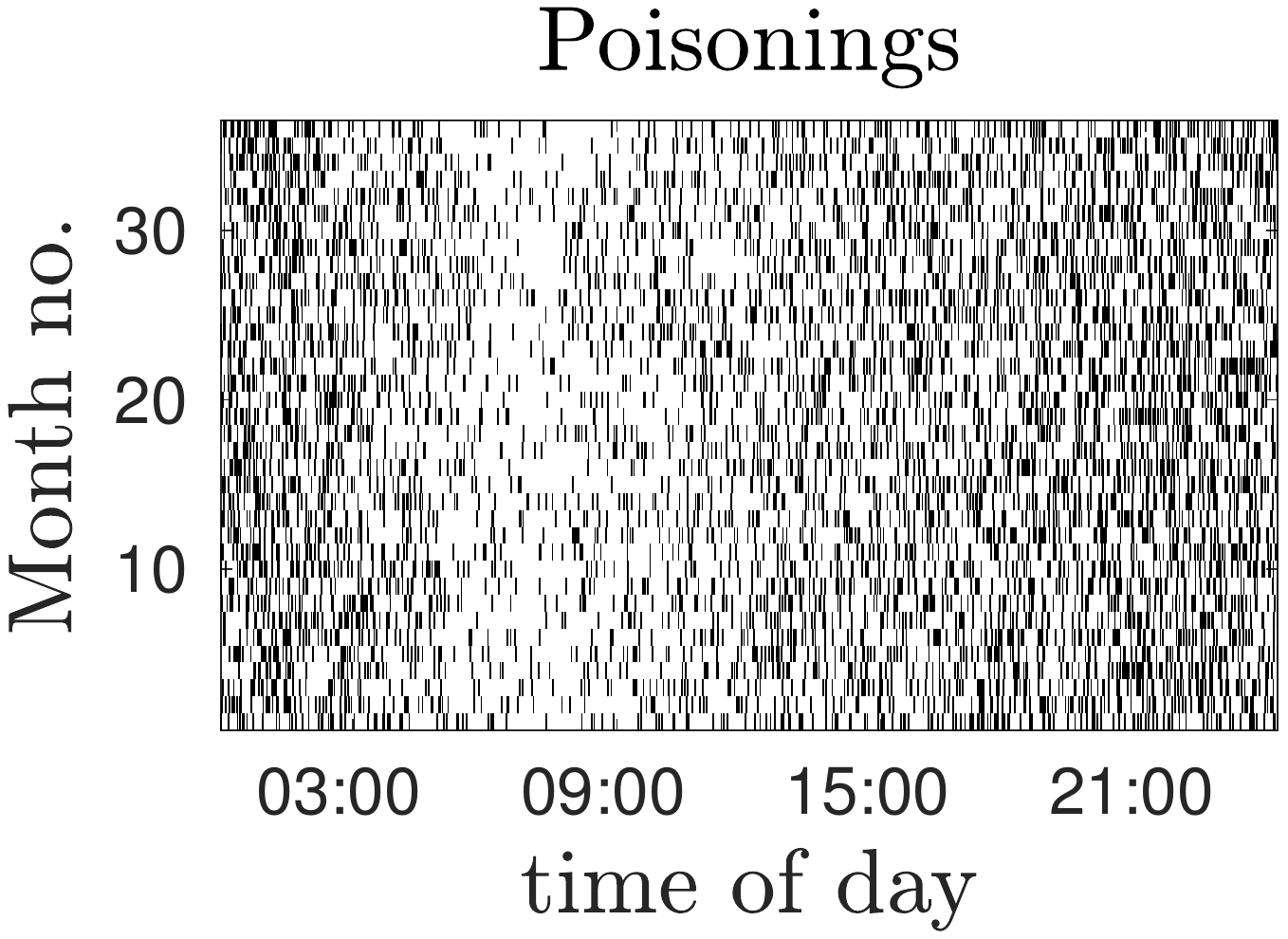}\hspace{0.5cm}
\includegraphics[width=0.32\textwidth]{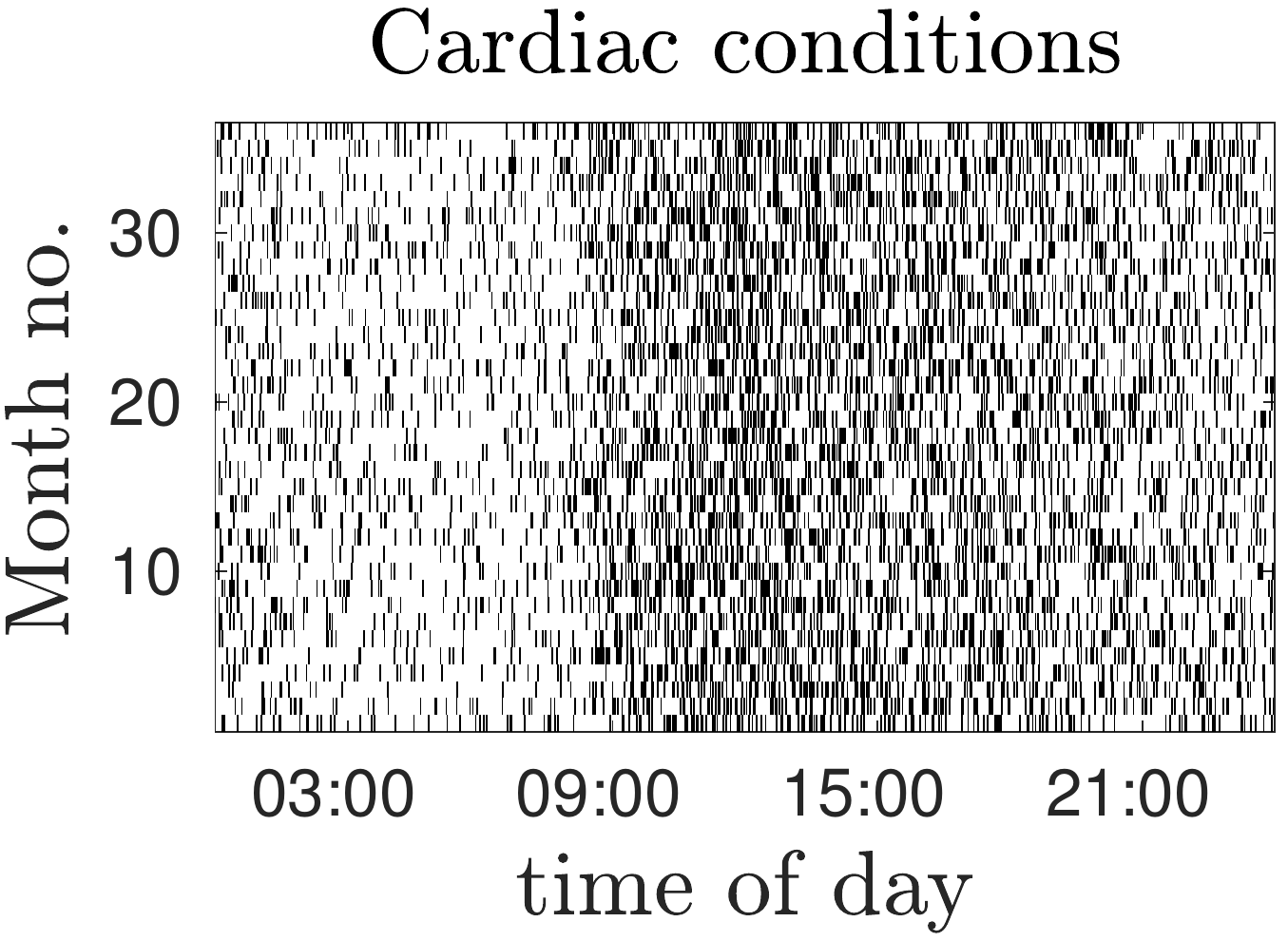}\\
 \textbf{Classification} \hspace{2cm} \textbf{Clustering}\\
\includegraphics[width=0.32\textwidth]{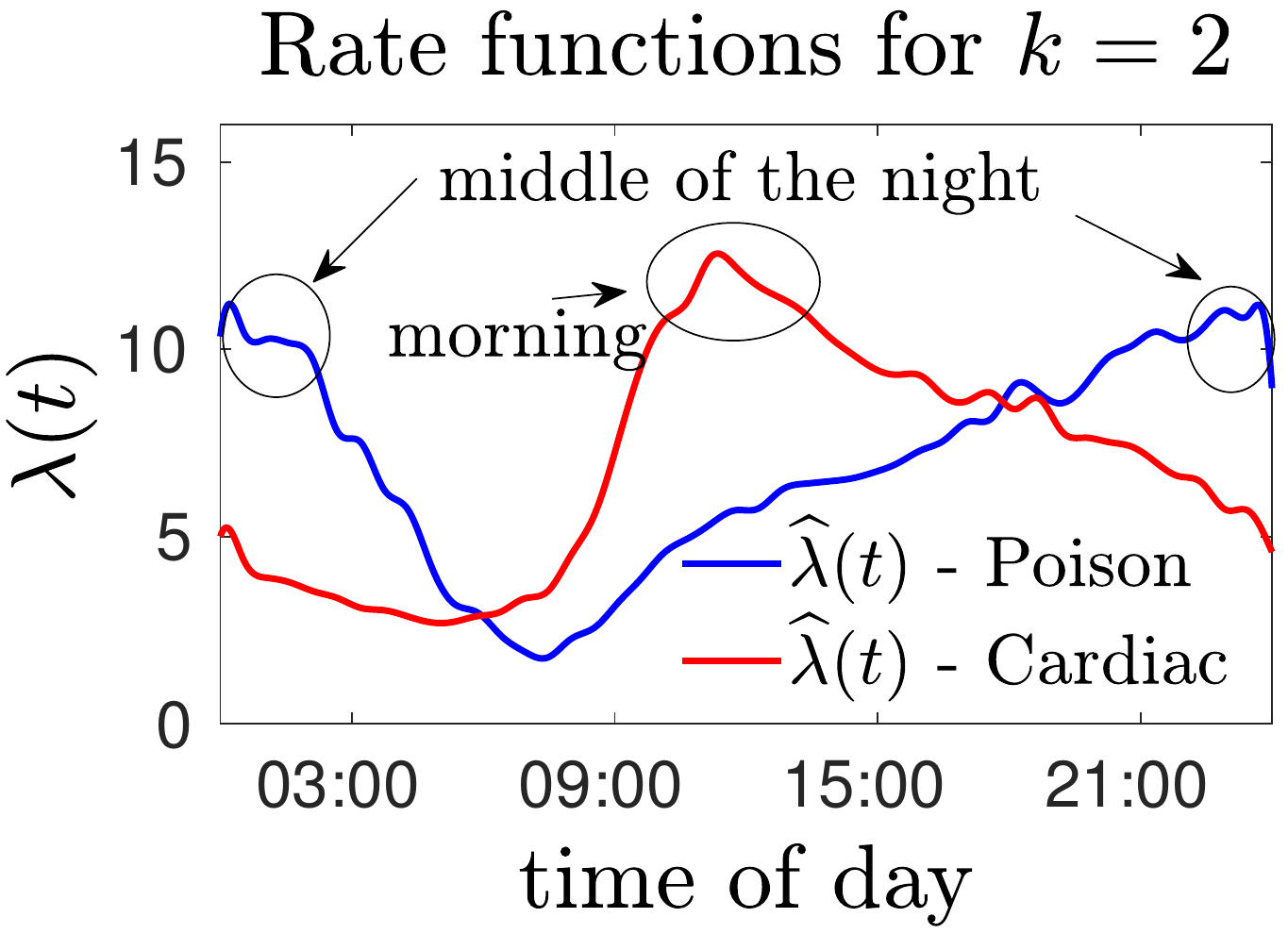}\hspace{0.5cm}
\includegraphics[width=0.32\textwidth]{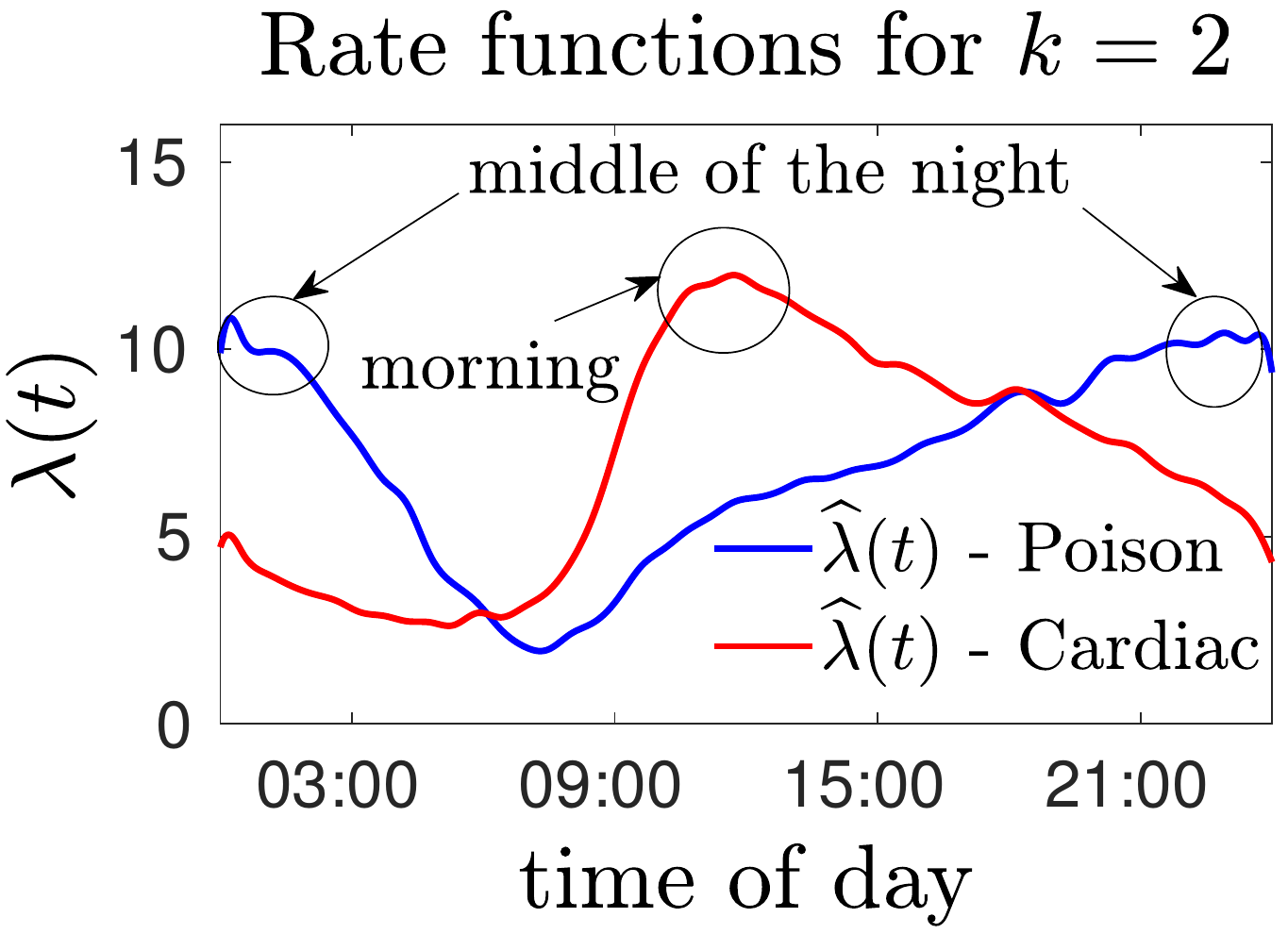}
\caption{Times of hospital admissions for a selection of observations as well as estimated rate functions for the A\&E diagnoses of poisoning (blue) and cardiac (red) obtained from our classification and clustering methods with $k=2$. The rate function estimates indicate that poisoning instances peak around midnight and cardiac conditions are maximal in the morning.}
\label{fig:AE}
\end{figure}

Estimated rate functions from the cluster analysis are also shown in Figure \ref{fig:AE} and these are very similar to the estimates obtained from the classification procedure. In particular, instances of poisoning peak around night time and cardiac conditions reach a high point in the morning. Assigning poisoning and cardiac months to the model for which their membership probabilities are maximal gives a clustering accuracy of 100\%. \\

We have made Matlab code for the classification and clustering procedures outlined in Sections \ref{sec:classification}-\ref{sec:clustering} and for the synthetic data set results in Section \ref{sec:synthetic_results} available at \noindent https://github.com/duncan-barrack/NHPP. The Hubway bike share data set is available at
\noindent https://www.thehubway.com/system-data. The store data and NHS A\&E data are not publicly available.  

\section{Discussion} \label{sec:conclusion}
In this paper, we have detailed principled approaches for the classification and clustering of observations of event data using NHPP models. Results on synthetic and real data were presented which show the effectiveness of our methods. The focus of this work has been on temporal point process data observed over a fixed interval $\{t: 0<t \leq T \}$. Another possibility, particularly suited for periodic data, would be to assume the domain of the data is periodic (e.g. a circle). The approach we have described here should generalise easily to the circular case. Furthermore, our method is not restricted to temporal data and, in principle, can be generalised to multi-dimensional spatial and spatial-temporal point process data. 

For the clustering procedure choosing the appropriate number of models in the NHPP mixture may not always be straightforward. In principle, the Bayesian information criterion \citep{schwarz1978estimating} or Akaike information criterion \citep{akaike1974new} could be used for this. Future work could investigate the suitability of these methods for choosing the number of NHPP components for our clustering procedure.

\section*{Acknowledgements}
We would like to thank The UK based retailer and NHS England for their support and for providing the data used in our analysis.  This work was funded by EPSRC grant EP/G065802/1 - Horizon: Digital Economy Hub at the University of Nottingham and EPSRC grant EP/L021080/1 - Neo-demographics: Opening Developing World Markets by Using Personal Data and Collaboration.

\bibliographystyle{plainnat}
\bibliography{NHPP}
\end{document}